\documentclass[a4paper,journal]{IEEEtran}
\usepackage{amsmath,amsfonts}
\usepackage{algorithm}
\usepackage{algpseudocode}
\usepackage{array}
\usepackage[caption=false,font=normalsize,labelfont=sf,textfont=sf]{subfig} 
\usepackage{textcomp}
\usepackage{stfloats}
\usepackage{url}
\usepackage{verbatim}
\usepackage{graphicx}
\usepackage{amssymb}
\usepackage[most]{tcolorbox}
\usepackage[T1]{fontenc}
\usepackage{multirow}%
\usepackage{booktabs}
\usepackage[sort&compress,numbers]{natbib}

\begin{document}

\title{REMoH: A Reflective Evolution of Multi-objective Heuristics approach via Large Language Models}

\author{
    \IEEEauthorblockN{
        Diego Forniés-Tabuenca\IEEEauthorrefmark{1}\IEEEauthorrefmark{2}, 
        Alejandro Uribe\IEEEauthorrefmark{1}\IEEEauthorrefmark{3}, 
        Urtzi Otamendi\IEEEauthorrefmark{2}\IEEEauthorrefmark{1},\\
        Arkaitz Artetxe\IEEEauthorrefmark{1}, 
        Juan Carlos Rivera\IEEEauthorrefmark{3},
        Oier Lopez de Lacalle\IEEEauthorrefmark{4}
    }
    
    \IEEEauthorblockA{\IEEEauthorrefmark{1}Vicomtech Foundation, Basque Research and Technology Alliance (BRTA), Donostia, Gipuzkoa, Spain} \\
    \IEEEauthorblockA{\IEEEauthorrefmark{2}Department of Computer Sciences and Artificial Intelligence, University of the Basque Country (UPV/EHU), Donostia, Gipuzkoa, Spain} \\
    \IEEEauthorblockA{\IEEEauthorrefmark{3}School of Applied Sciences and Engineering, Universidad EAFIT, Medellin, Antioquia, Colombia}\\
    \IEEEauthorblockA{\IEEEauthorrefmark{4}HiTZ Basque Center for Language Technology, University of the Basque Country (UPV/EHU), Donostia, Gipuzkoa, Spain}
    \thanks{Corresponding author: Diego Forniés-Tabuenca}}

\markboth{Pre-print submitted to IEEE Transactions on Evolutionary Computation}%
{Fornies \MakeLowercase{\textit{et al.}}: REMoH:  A Reflective Evolution of Multi-objective Heuristics approach via Large Language Models}

\maketitle

\begin{abstract}
Multi-objective optimization is fundamental in complex decision-making tasks. Traditional algorithms, while effective, often demand extensive problem-specific modeling and struggle to adapt to nonlinear structures. Recent advances in Large Language Models (LLMs) offer enhanced explainability, adaptability, and reasoning. This work proposes Reflective Evolution of Multi-objective Heuristics (REMoH), a novel framework integrating NSGA-II with LLM-based heuristic generation. A key innovation is a reflection mechanism that uses clustering and search-space reflection to guide the creation of diverse, high-quality heuristics, improving convergence and maintaining solution diversity. The approach is evaluated on the Flexible Job Shop Scheduling Problem (FJSSP) in-depth benchmarking against state-of-the-art methods using three instance datasets: Dauzere, Barnes, and Brandimarte. Results demonstrate that REMoH achieves competitive results compared to state-of-the-art approaches with reduced modeling effort and enhanced adaptability. These findings underscore the potential of LLMs to augment traditional optimization, offering greater flexibility, interpretability, and robustness in multi-objective scenarios.
\end{abstract}


\section{Introduction}
\IEEEPARstart{T}{he} core challenge in many domains lies in optimization: the quest to find the best solution from a set of available alternatives. Optimization algorithms are essential tools for this quest across diverse fields, including scheduling, logistics, finance, healthcare, and artificial intelligence \cite{HUANG2024101663}.  As data complexity and scale increase, the demand for a efficient optimization strategies becomes more critical \cite{gad2022particle}. While traditional optimization algorithms have achieved notable success, their design often relies heavily on domain-specific expertise. This reliance can limit adaptability and scalability, making responding to dynamic or poorly defined problem structures difficult \cite{tang2021review}.


Large Language Models (LLMs) emerge as transformative tools in the field of optimization and decision-making, offering distinct advantages over traditional methods. Unlike conventional approaches such as metaheuristics or Reinforcement Learning (RL), which often suffer from lengthy training phases and convergence to local optima, LLMs utilize pre-trained experiential knowledge to enable real-time, self-adaptive scheduling \cite{LIU2025110889}. This capability positions LLMs not only as optimization tools but also as communicative agents that align production goals with dynamic decision-making requirements. However, the practical deployment of LLMs in engineering contexts faces some challenges. Their inherent biases and potential for factual inaccuracies necessitate robust grounding mechanisms. Current implementations of LLMs, while capable of interpreting and executing textual scheduling commands, often lack direct alignment with the physical and operational realities of manufacturing environments \cite{GKOURNELOS20249}. This highlights the critical role of self-supervised learning and other calibration strategies to ensure the reliability and applicability of LLMs in industrial settings.

A promising direction involves the synergistic integration of LLMs with classical optimization algorithms. This hybridization leverages LLMs’ extensive domain knowledge to enrich modeling and strategic decision-making, while optimization algorithms enhance LLM behavior and output quality. Such combinations pave the way for intelligent search operations, adaptive algorithm configuration, and even the autonomous generation of new optimization strategies, thereby reducing the dependence on human expertise and expanding the scope of problem-solving in complex domains \cite{HUANG2024101663}. Evolutionary algorithms (EAs), in particular, present complementary strengths when paired with LLMs. While EAs offer robust global search mechanisms in black-box optimization contexts, LLMs contribute by guiding these searches with contextual knowledge and textual reasoning capabilities. This mutual enhancement broadens the range of problems each technique can effectively address and fosters more intelligent and efficient exploration in diverse computational environments \cite{wu2024evolutionarycomputationeralarge}.

Recent advancements further demonstrate the potential of LLMs to automate the generation and evolution of optimization components such as heuristics, search operators, and algorithmic frameworks \cite{Pluhacek2023, HUANG2024101663, ZHONG2025103042}. Through their natural language understanding and generative abilities—especially in code synthesis—LLMs can function as optimizers or co-designers in evolutionary processes. These novel approaches aim to reduce manual intervention, facilitate intelligent adaptation to specific problem instances, and enhance exploration-exploitation trade-offs \cite{liu2024evolutionheuristicsefficientautomatic}. Despite significant progress in evolutionary and mathematical optimization techniques, designing adaptive and generalizable heuristics for multiple objectives remains challenging.

The contribution of this paper is a novel framework that leverages LLMs to enhance multi-objective optimization through dynamic heuristic generation and search space reflection. Specifically, we introduce \textit{Reflective Evolution of Multi-objective Heuristics} (REMoH), a hybrid framework designed to evolve high-quality and diverse heuristic operators for multi-objective problems. This paper proposes the following key contributions:

\begin{itemize}
    \item \textbf{REMoH Framework:} A novel optimization methodology that integrates LLM-generated heuristic operators within an NSGA-II selection process. The LLM generates domain-agnostic, human-readable heuristics, improving explainability and adaptability across problem instances.

    \item \textbf{Reflection Mechanism:} An innovative component that performs clustering on the current heuristic population and uses the reflection mechanism to improve search space exploration, prevent premature convergence, and enhance the diversity of solutions.

    \item \textbf{Ablation Study:} A detailed ablation analysis demonstrates the effectiveness of the reflection mechanism. Results show significant gains in Hypervolume (HV) and Inverted Generational Distance (IGD) when the reflection mechanism is included.

    \item \textbf{Comprehensive Benchmarking on FJSSP:} The proposed approach is benchmarked against mathematical modeling (Mixed-Integer Linear Programming and Constraint Programming) and state-of-the-art learning-based (Reinforcement Learning) methods using the Flexible Job Shop Scheduling Problem (FJSSP). Experiments are conducted on the widely used Brandimarte suite.
    
    \item \textbf{Modeling Flexibility and Constraint Integration:}  Unlike traditional mathematical approaches that struggle with non-linear constraints, the proposed framework can integrate complex and context-sensitive constraints with little reformulation effort, demonstrating robustness and flexibility.
\end{itemize}

The remainder of this article is organized as follows:
Section \ref{sec: sota} reviews related work on integrating LLMs in multi-objective optimization.
Section \ref{sec: method} presents the proposed REMoH framework in detail.
Section \ref{sec:experimental-setup} outlines the experimental setup, including datasets, evaluation metrics, and baseline methods.
Section \ref{sec: results} discusses the experimental results, highlighting both the benchmarking analysis and the ablation study.
Finally, Section \ref{sec:Conclusions} concludes the paper and suggests directions for future research.

\section{State of the art}
\label{sec: sota}

Building upon the foundational capabilities of LLMs in optimization and decision-making, recent research has expanded their utility through integration with other artificial intelligence methodologies. Notably, the combination of RL and LLMs has demonstrated significant promise in decision support systems, particularly in domains such as crop management. Here, RL contributes adaptive policy learning, while LLMs provide semantic context and domain-specific knowledge, enabling more nuanced and context-aware decisions \cite{CHEN2025110028}. A concrete application involved integrating deep Q-networks (DQN), the DistilBERT model, and crop simulations to generate and interpret state variables. These were translated into language representations to enhance the LLM's interpretive capacity, leading to substantial economic performance gains in real-world agricultural settings. However, in this case, the LLM functioned primarily as a state encoder, indicating substantial untapped potential for deeper LLM integration beyond auxiliary roles in RL frameworks \cite{wu2024newagronomistslanguagemodels}.

In parallel, novel systems like AutoML-GPT illustrate the growing relevance of LLMs in automating complex machine learning pipelines. By combining GPT-based models with dynamic hyperparameter optimization, AutoML-GPT can autonomously process user requirements, generate suitable prompts, select model architectures, tune parameters, and manage the entire experimental lifecycle \cite{zhang2023automlgptautomaticmachinelearning}. Other approaches have investigated the potential of LLMs in hyperparameter optimization (HPO). These approaches move beyond conventional search spaces: some allow the LLM to identify optimizable parameters and their bounds autonomously, and even to treat elements such as model code as a hyperparameter, vastly expanding the HPO design space \cite{zhang2024usinglargelanguagemodels}. Architectures like AgentHPO implement multi-agent systems in which an LLM-based Creator agent simulates expert reasoning to propose high-quality hyperparameter configurations, followed by an Executor agent that evaluates them, streamlining a process typically dependent on substantial domain knowledge and computational resources \cite{liu2025largelanguagemodelagent}.

Further refinement of LLMs' reasoning capacity has been achieved through advanced prompting techniques, particularly chain-of-thought (CoT) prompting. Unlike standard few-shot learning, CoT prompting supplies LLMs with intermediate reasoning steps, enabling them to decompose complex tasks into structured sub-problems that guide inference processes more effectively \cite{LIU2025129190}. Recent developments, such as the PromptBreeder system, have taken this a step further by employing CoT prompting in a self-referential, evolutionary framework. PromptBreeder iteratively improves both task and mutation prompts, allowing LLMs to self-optimize over time. This approach has demonstrated superior performance over standard prompting strategies, enhancing LLM effectiveness in domains requiring sophisticated reasoning and interpretability \cite{fernando2023promptbreederselfreferentialselfimprovementprompt}.

Beyond their integration with external AI techniques and advanced prompting strategies, recent developments have increasingly focused on using LLMs as intrinsic optimizers. One notable contribution is Optimization by PROmpting (OPRO), a technique where the optimization problem is formulated in natural language, and the LLM iteratively proposes new candidate solutions based on prior feedback embedded within the prompt \cite{yang2024largelanguagemodelsoptimizers}. Empirical validation on tasks such as linear regression and the Traveling Salesman Problem (TSP) demonstrated substantial performance gains, underscoring the potential of prompt-based optimization without explicit mathematical modeling. Parallel work has explored the direct formulation and solution of mathematical programming problems using LLMs. The OptiMUS framework, for instance, leverages LLMs to interpret natural language problem descriptions and translate them into structured Mixed-Integer Linear Programming (MILP) formulations \cite{ahmaditeshnizi2023optimusoptimizationmodelingusing}. Beyond model generation, this approach is capable of debugging code, developing test cases, and verifying solutions, illustrating a comprehensive pipeline for language-driven optimization. Other techniques employ neural bandit algorithms for black-box optimization of prompts balancing exploration and exploitation without relying on traditional Bayesian optimization techniques \cite{lin2024useinstinctinstructionoptimization}.

An emerging line of research leverages the generative and reasoning capacities of LLMs to design novel heuristics and metaheuristic algorithms, advancing the field of algorithmic optimization beyond traditional boundaries. For instance, the LLM assisted Hyper-heuristic Optimization Algorithm (LLMOA) integrates LLMs as high-level controllers that, through prompt engineering, generate adaptive optimization strategies. This architecture is complemented by low-level elite-based local search heuristics —grounded in the proximate optimality principle— and enriched with operators from differential evolution, producing a diverse and effective set of search behaviors \cite{ZHONG2025103042}. Similarly, GPT-4 has been employed to synthesize hybrid optimization algorithms by combining characteristics of well-known metaheuristics such as cuckoo search, whale optimization, particle swarm optimization, grey wolf optimizer, and others \cite{Pluhacek2023}. Further exploration into LLM-aided algorithm design has led to the development of entirely new metaheuristics, such as the zoological search optimization (ZSO), created through ChatGPT's text generation capabilities guided by the structured CRISPE prompting framework \cite{Zhong2024-tr}.

Recent advances have demonstrated the potential of integrating LLMs within evolutionary heuristic frameworks, fostering a new class of adaptive optimization methods. This synergistic interaction between evolutionary algorithms (EAs) and LLMs is gaining recognition as a promising direction for tackling generative and optimization tasks, with LLMs enriching the intelligence of EAs and EAs offering global search mechanisms for optimizing LLM-driven processes \cite{10767756}. This bidirectional influence is shaping a novel paradigm in optimization research, merging the strengths of both paradigms toward more general and autonomous problem-solving capabilities.

One important research direction involves the evolutionary generation of optimization algorithms themselves, where LLMs are embedded within the evolutionary loop to produce, mutate, and refine algorithms across iterations. For instance, the Algorithmic Evolution using LLMs (AEL) approach \cite{liu2023algorithmevolutionusinglarge} interacts with LLMs as part of an evolutionary cycle to automatically generate constructive algorithms tailored to specific problems, outperforming both handcrafted and purely LLM-generated baselines. Expanding on this concept, a similar framework \cite{liu2024evolutionheuristicsefficientautomatic} was applied to design a Guided Local Search algorithm for the TSP, demonstrating the practicality of fully automating algorithm creation without prior model training. In a complementary effort, LLaMEA \cite{vanstein2024} presents a framework where LLMs iteratively generate, mutate, and evaluate optimization algorithms based on task descriptions and performance feedback, offering an efficient and expert-free route to generate novel metaheuristics. Likewise, LLMs-GP \cite{hemberg2024evolvingcodelargelanguage} adapts Genetic Programming principles by leveraging LLMs' pattern recognition capabilities to design genetic operators and produce executable optimization code. A more dynamic variant, CoCoEvo \cite{li2025cocoevocoevolutionprogramstest}, proposes a co-evolutionary framework where both programs and test cases evolved simultaneously from natural language task descriptions, removing the dependency on pre-defined tests and employing LLM-based operators to drive the co-evolution process.

Another line of work focuses on using LLMs as evolutionary or heuristic operators to improve solution generation and enhance optimization quality. LEO \cite{brahmachary2024largelanguagemodelbasedevolutionary}, the Language-model-based Evolutionary Optimizer, incorporates LLMs in a population-based strategy with separate explore and exploit pools, enabling robust search performance in both single and multi-objective settings. A further contribution in this direction is the LLM-driven Evolutionary Algorithm (LMEA), which introduces a lightweight but effective paradigm where LLMs autonomously conduct selection and genetic operations without requiring retraining or expert input \cite{liu2024largelanguagemodelsevolutionary}. Notably, it integrates a self-adaptive temperature mechanism that dynamically balances exploration and exploitation, achieving competitive results compared with traditional heuristics on benchmark combinatorial tasks like the TSP. Similarly, QDAIF \cite{bradley2023qualitydiversityaifeedback} introduces a quality-diversity framework where LLMs are used not only to generate candidate solutions but also to evaluate their diversity and quality, facilitating a more informed and iterative evolutionary process. In terms of specific operator design, LMX \cite{meyerson2024languagemodelcrossovervariation} and difference modeling \cite{Bradley2024} offer mechanisms where LLMs function as crossover operators by synthesizing offspring from multiple parents, or by predicting code differences to model solution variation. In the multi-objective domain, researchers \cite{liu2024largelanguagemodelmultiobjective} have proposed using LLMs as black-box search operators within decomposed subproblems of a multi-objective optimization problem, enabling zero-shot optimization without problem-specific learning or expert tuning.

Beyond algorithm and solution generation, several works explore LLMs for heuristic design and adaptation, especially in scheduling and combinatorial optimization. SeEvo \cite{huang2024automaticprogramminglargelanguage} introduces a self-evolutionary population model where LLMs not only generate dispatching rules for dynamic job shop scheduling problems but also evolve them over time, inspired in human-like reflection mechanisms, to guide both individual creation and population exploration. Similarly, a two-phase task scheduling framework \cite{yatong2024tseohedgeservertask} leverages LLMs to first generate heuristic strategies and corresponding code and then evaluate and evolve these scheduling schemes across generations. A broader formulation is provided by the Language Hyper-Heuristics (LHHs) framework \cite{ye2024reevolargelanguagemodels}, where LLMs generate heuristics with minimal human input, and are enhanced by Reflective Evolution (ReEvo), a method combining verbal feedback from LLMs with evolutionary search to explore and refine large spaces of possible heuristic combinations. Finally, the LLM-EPS framework \cite{dat2024hsevoelevatingautomaticheuristic} addresses the persistent challenge of balancing exploration and exploitation in heuristic search spaces by introducing diversity metrics inspired by Shannon Entropy. To support this, HSEvo, a variant of LLM-EPS, incorporates harmony search principles and low-cost LLM interactions ("flash reflection") to promote diversity while improving convergence efficiency.

In contrast to existing approaches, this research introduces a novel framework that integrates LLM-based heuristic generation with a reflective mechanism grounded in clustering and performance introspection. Unlike prior efforts that either use LLMs passively to generate heuristics or evolve them without structured feedback, REMoH actively guides the evolutionary process through dynamic, multi-level reflection. This enables the adaptive generation of diverse, high-performing multi-objective heuristics with minimal human intervention, addressing the open challenge of maintaining both exploration and convergence in heuristic evolution for complex scheduling problems.

\section{Methodology}
\label{sec: method}

\begin{figure*}[ht]
  \centering
  \includegraphics[width=0.8\textwidth]{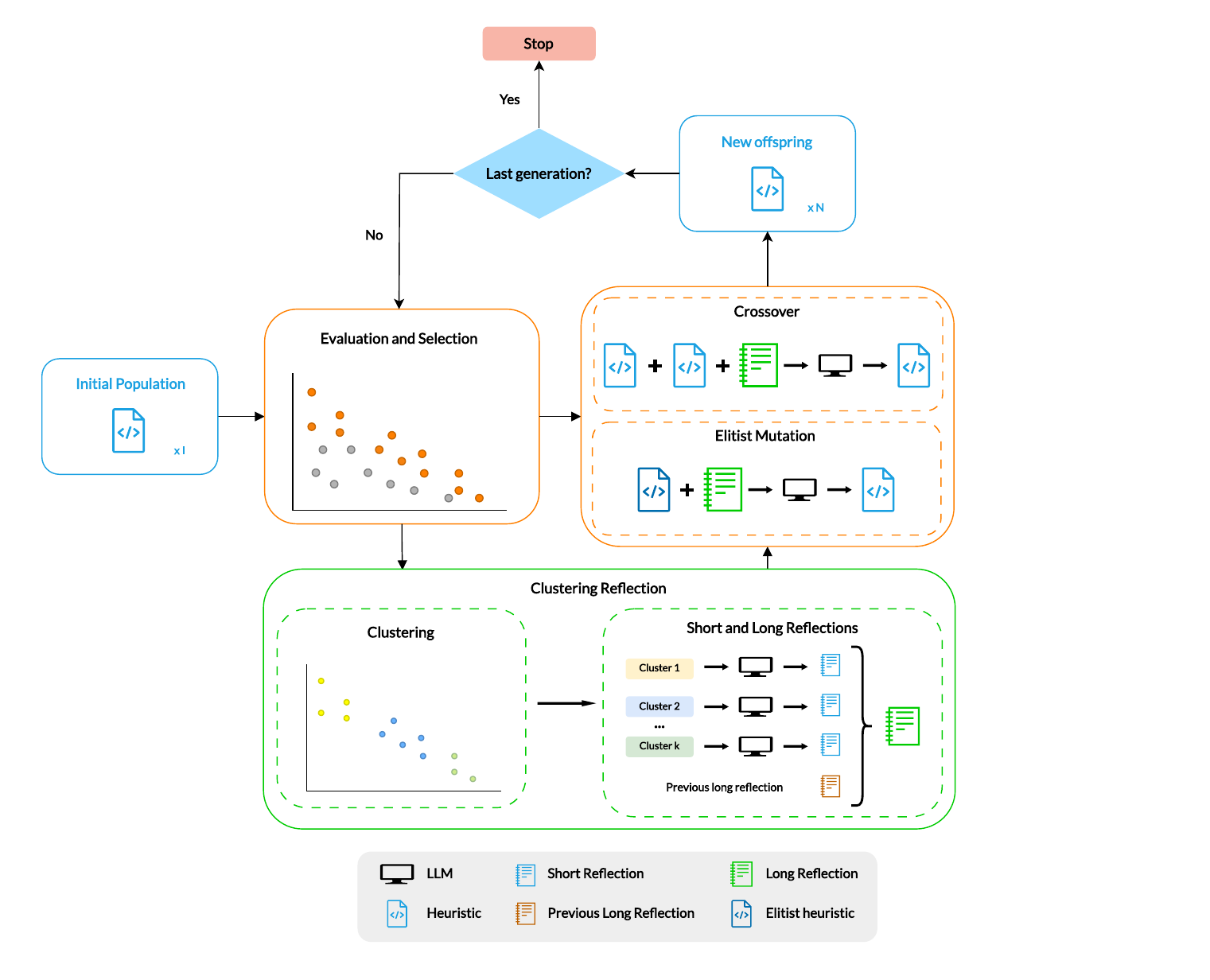}
  \caption{Detailed diagram of our proposed Multi-Objective Evolution of Heuristics-based optimization methodology.}
  \label{fig:algorithmSchema}
\end{figure*}

In this section, we introduce REMoH, a novel algorithm that integrates LLMs with the traditional NSGA-II framework to generate heuristics for solving multi-objective optimization problems. Additionally, we introduce an innovative reflection mechanism to assist the algorithm in discovering new heuristics that efficiently dominate the Pareto front, aiming for its convergence toward the optimal front. 

As shown in Figure \ref{fig:algorithmSchema}, the process begins with the initialization of the population using dynamic prompts to foster diversity. Each heuristic is evaluated and selected according to the NSGA-II framework, based on Pareto dominance and crowding distance. Next, a clustering-based reflection is performed to extract key features from each group of heuristics, identifying their strengths and weaknesses to guide the next stage. In this phase, new offspring are generated through reflective crossover and elitist mutation. Finally, the newly generated heuristics are evaluated and selected alongside their parents, or the process terminates if the final generation has been reached.

The following sections provide a detailed description of each step in the process.

\textbf{Individual encoding}. Following the principles of LLM-based Evolutionary Program Search (LLM-EPS) approaches like EoH  \cite{liu2024evolutionheuristicsefficientautomatic} and ReEvo \cite{ye2024reevolargelanguagemodels}, REMoH represents each individual as a code snippet generated by an LLM. This encoding approach allows for high flexibility, as individuals are not constrained by predefined structures, and enables direct evaluation within the problem context. By evolving code rather than fixed parameters, REMoH dynamically adapts its heuristic strategies, making it well-suited for complex scheduling and optimization tasks.

\textbf{Population initialization}. REMoH initializes each heuristic by providing the LLM with the task specification. The task specification includes the description of the optimization problem and the function details, outlining both the input and output formats. To leverage existing knowledge, the framework can be initialized with a seed heuristic function. To help the model diversify the initial population, we have drawn on the HSEvo strategy \cite{dat2024hsevoelevatingautomaticheuristic}, which alters the role instruction to generate different prompts. This approach aids the model in enriching its initial heuristics. Typically, the initial population is larger than in subsequent generations. This is a common strategy in evolutionary algorithms to enhance diversity in the first generation.

\textbf{Evaluation}. To assess the performance of the heuristics, they are tested across various problem instances, with each heuristic receiving a score for every objective function. These scores are then normalized using Standard Scaler and averaged to evaluate overall performance. Outlier scores do not contribute to the normalization parameters to prevent bias from extremes. Furthermore, if a heuristic is infeasible or contains implementation errors, the LLM is prompted to generate a new one to prevent information loss.

\textbf{Selection}. Based on the evaluation, our algorithm selects parents using the NSGA-II approach, ranking individuals based on their dominance against others. Then, for individuals in the last front to be selected, we choose them based on their crowding distance, ensuring diversity within the population.

\textbf{Clustering Reflection}. The LLM generates reflection through clustering parents based on their performance regarding dominance. Our reflection contains three main steps:

\begin{itemize}
    \item Group the parents based on their fitness in different objective functions using K-Means clustering. The number of clusters (K) is determined through the silhouette method due to its performance when used in multi-objective algorithms \cite{Gavval2020}.
    \item Generate a brief reflection for each cluster by prompting the LLM with the code from each individual and their overall performance based on the cluster centroid. This reflection will summarize the heuristics, emphasizing their main characteristics and key features.
    \item Produce the final reflection providing the LLM with all short cluster reflections, their general performance, and previous long-term reflections. In this final step, the goal is to analyze the strengths and weaknesses of each group. The LLM identifies emerging trends and proposes potential directions for future exploration.
\end{itemize}

This clustering reflection process ensures a strategic balance between exploration and exploitation. By synthesizing performance patterns and heuristic structures across different objective clusters, the LLM guides the generation of more informed and effective offspring capable of tackling multi-objective optimization.

\textbf{Crossover}. In this stage, new offspring algorithms are created by merging components from the parent algorithms. The main objective is to generate new heuristics based on the parents, leveraging their strengths and avoiding their weaknesses, as determined by the clustering reflection. The LLM is provided with the crossover task specifications, including the two parents' code and the reflection.

\textbf{Elitist Mutation}. This step involves selecting the individual with the best fitness achieved so far, defined as the one obtaining the highest average across all objective functions. This criterion promotes solutions that achieve a balanced trade-off among objectives. Following the HSEvo framework \cite{dat2024hsevoelevatingautomaticheuristic}, we incorporate reflection into this mutation process, ensuring that the improvement is not purely random but guided by prior analysis. The LLM is guided by mutation task specifications, comprising the elite individual's code and its reflection.

\section{Experimental Setup}
\label{sec:experimental-setup}

This section describes the experimental setup used to evaluate the proposed REMoH algorithm. The aim is to assess the effectiveness of our multi-objective optimization framework through a well-established and practically relevant problem: the FJSSP \cite{xiong2022survey}.

Scheduling problems are critical in industrial and operational environments since they directly impact efficiency, resource utilization, and overall productivity \cite{DAUZEREPERES2024409, xie2019review}. Among these, the Job Shop Scheduling Problem (JSSP) is particularly challenging due to its combinatorial nature and the complexity of assigning operations to machines while optimizing objectives. The generalization of JSSP that considers a set of eligible machines that can process each operation is named FJSSP. This additional flexibility increases the complexity and applicability of the problem in real-world manufacturing systems \cite{leusin2018solving}.

The main goal in FJSSP is to determine the optimal assignment of operations to machines and the sequencing of operations on each machine, to optimize objectives such as minimizing the makespan (total completion time), balancing machine workloads, or minimizing total idle time.

This experimentation phase focuses on two specific objectives: (1) \textit{Makespan}, defined as the total time required to complete all scheduled jobs, and (2) \textit{Workload Balance}, which measures the maximum workload across all machines. These are widely recognized key performance indicators (KPIs) in the scheduling literature. While makespan minimization is traditionally the primary focus, we include workload balance as an additional objective to evaluate the performance of the multi-objective framework under a broader set of constraints.

\subsection{Evaluation and Selection Criteria}
\label{subsec: eval}

This section describes the evaluation methodology used to compare and select the most effective heuristics within the proposed framework. As introduced in the methodology section, each heuristic is evaluated by solving all problem instances in the dataset. For each instance, the objective values are normalized across the population, and the mean of these normalized values is computed per objective to represent overall performance.

In the final evaluation, a similar procedure is applied across all instances and generations. Specifically, the initial population and each generation are evaluated on the full set of FJSSP instances. The resulting objective values are normalized per instance using a robust normalization strategy based on percentiles to reduce the influence of outliers. This normalization is applied consistently across all independent experiments being compared, ensuring that the scores are scaled relative to overall performance and can be fairly evaluated. After normalization, mean objective values are computed for each individual across all instances.

To evaluate the quality of the obtained heuristics and the resulting Pareto front across generations, two standard multi-objective optimization metrics are employed:

\begin{itemize}
    \item \textbf{Hypervolume (HV)}: It measures the volume of the objective space dominated by the obtained Pareto front with respect to a reference point, effectively capturing both convergence and diversity. After a calibration procedure, the reference point was set at 0.3 for each objective function axis.
    \item \textbf{Inverted Generational Distance (IGD)}: Calculates the average distance from points in a reference Pareto front to the nearest point in the obtained front, providing insight into convergence and distribution.
\end{itemize}

Due to the lack of ground truth Pareto fronts in most FJSSP datasets, the reference set is constructed as the non-dominated front obtained from the union of all solutions generated by competing heuristics. Then, this \textit{Pareto} is used to compute the metrics for all individuals.

Additionally, the average IGD and HV values are computed per generation to evaluate the evolution of heuristic quality over generations. This helps analyze the convergence and diversity preservation throughout the evolutionary process. This evaluation procedure is applied consistently across experiments, including the benchmarking and ablation studies discussed in the results section.

Also, to avoid the stochastic nature of evolutionary algorithms, each metric is obtained by averaging the results obtained in three independent runs, ensuring statistical robustness.

\subsection{Instances}

For strict evaluation and to facilitate comparative benchmarking of the proposed methodology, we employed three established and publicly available benchmark datasets relevant to the FJSS problem:

\begin{itemize}
    \item \textbf{Brandimarte:} In his seminal work \textit{Routing and scheduling in a FJSSP by tabu search}, Brandimarte et al \cite{Brandimarte1993} introduced a set of instances labeled MK01 to MK15. These vary in size and are primarily designed for evaluating approaches focused on makespan minimization, though they are adaptable to other objectives.
    
    \item \textbf{Barnes:} One of the earliest public datasets for the FJSSP, introduced by Barnes et al. \cite{BARNES01041995}, alongside a novel tabu search approach. These instances remain a foundational benchmark in the literature.

    \item \textbf{Dauzere:} This dataset \cite{DAUZEREPERES2024409} consists of 18 problem instances varying in size and flexibility. It is a widely adopted benchmark for evaluating FJSSP algorithms and is critical for assessing performance and scalability in newly developed scheduling techniques.
\end{itemize}

Among these, the Brandimarte dataset has been widely utilized in numerous studies for benchmarking heuristic and metaheuristic algorithms in the FJSSP \cite{DAUZEREPERES2024409}. For this reason, it was selected for in-depth evaluation of solution quality and robustness. Specifically, in the Ablation Study \ref{subsec:AblationStudy} and Flexibility Evaluation \ref{subsec:robustness-evaluation}, Brandimarte was used solely for training, since these experiments focused on model behavior and did not require external validation. Conversely, the other datasets were used for training in the Benchmarking Methods \ref{subsec:brandimarte-benchmark}. At the same time, Brandimarte served as the evaluation set, enabling a fair comparison of generalization and performance on a widely accepted benchmark.

\subsection{Benchmarking Methods}
\label{subsec:otherMethods}

The experimentation process includes a comprehensive benchmarking analysis against classical optimization methods and recent state-of-the-art learning-based techniques. This comparison aims to assess not only solution quality but also flexibility, scalability, and adaptability of our approach across a diverse set of FJSSP instances.

Following the work of Dauzere et al. \cite{DAUZEREPERES2024409}, exact methods such as Branch and Bound (B\&B), Mixed-Integer Linear Programming (MILP), and Constraint Programming (CP) have historically been employed to tackle the FJSSP. In this sense, CP has demonstrated superior performance, particularly in solving medium-sized instances optimally and generating feasible solutions for larger and more complex scenarios. MILP, while offering an exact mathematical representation of scheduling constraints and objectives, has challenges with problem size due to its combinatorial nature.

In the benchmarking framework, we include three classical optimization methods and two recent deep learning-based approaches inspired by RL. Including RL methods reflects a growing interest in data-driven, adaptive optimization strategies capable of generalizing across problem instances. These learning-based methods generate learning policies that directly map states to scheduling actions, potentially offering better scalability and adaptability in dynamic environments.

Additionally, we employ a lexicographic optimization strategy to ensure a fair comparison across methods, especially those that do not natively support multi-objective optimization. In this strategy, one solution is obtained by prioritizing \textit{makespan} as the primary objective and \textit{workload balance} as the secondary; a second solution is derived by reversing the priorities, \textit{workload balance} first, then \textit{makespan}. This process yields two points: in many cases, they coincide, indicating aligned optima, while in others, they differ, capturing the trade-off between objectives. This dual evaluation enables consistent comparison across single- and multi-objective solvers.

\subsubsection{Mixed Integer-Linear Programming}
\label{subsec:MILP}

This section presents the MILP formulation for the FJSSP, based on the model presented by Dauzere et al. \cite{DAUZEREPERES2024409}. The formulation captures the essential constraints of FJSSP, ensuring a feasible and optimal assignment of operations to machines while minimizing the makespan or workload balance. The lexicographic process allows us to construct two points of the Pareto front, providing a strong baseline to compare against the results obtained by the algorithms developed in our proposed methodology. The following describes the notations used, sets, parameters, variables, and the formulation.

\begin{table}[ht]
\centering
\caption{Summary of Sets, Parameters, and Variables}
\begin{tabular}{c|p{0.8\linewidth}}
\hline
\multicolumn{2}{c}{\textit{Sets}} \\
\hline
$O$ & Set of operations \\
$pr(i)$ & Predecessor of operation $i$ \\
$R$ & Set of machines \\
$R_i$ & Subset of machines $R$ capable of performing operation $i$ \\
\hline
\multicolumn{2}{c}{\textit{Parameters}} \\
\hline
$p_{k,i}$ & Processing time of operation $i$ on machine $k$ \\
$H$ & A sufficiently large number (big-M constant) \\
$C$ & Fixed makespan \\
$RM$ & Fixed maximum machine operative time \\
\hline
\multicolumn{2}{c}{\textit{Variables}} \\
\hline
$\alpha_{k,i}$ & Binary variable: 1 if operation $i$ is assigned to machine $k$ \\
$\beta_{i,i'}$ & Binary variable enforcing sequencing between operations on the same machine \\
$t_i$ & Start time of operation $i$ \\
$c_{max}$ & Resulting makespan \\
$r_{max}$ & Maximum machine operative time (used for workload balance) \\
\hline
\end{tabular}
\label{tab:notation}
\end{table}

\begin{figure*}[ht]
\begin{align}
& \min \ c_{max}& \label{eq1} \\
& \sum_{k \in R_i} \alpha_{k,i} = 1 & \quad \forall \ i \in O \label{eq2}\\
& t_i \geq t_{pr(i)} + \sum_{k \in R_{pr(i)}} p_{k,pr(i)} \cdot \alpha_{k,pr(i)} & \quad \forall \ i \in O \label{eq3}\\
& t_i \geq t_{i'} + p_{k,i'} - (2-\alpha_{k,i}-\alpha_{k,i'}+\beta_{i,i'}) \cdot H & \quad \forall \ (i,i') \in O \times O, \ s.t \ i \neq i', \ \forall \ k \in R_i \cap R_{i'} \label{eq4}\\
& t_{i'} \geq t_{i} + p_{k,i} - (3-\alpha_{k,i}-\alpha_{k,i'}+\beta_{i,i'}) \cdot H & \quad \forall \ (i,i') \in O \times O, \ s.t \ i \neq i', \ \forall \ k \in R_i \cap R_{i'} \label{eq5}\\
& c_{max} \geq t_{i} + \sum_{k \in R_{i}} p_{k,i} \cdot \alpha_{k,i} & \quad \forall \ i \in O \label{eq6}\\
& \alpha_{k,i} \in \{ 0,1 \} & \quad \forall \ i \in O, \ k \in R_i \label{eq7}\\
& \beta_{i,i'} \in \{ 0,1 \} & \quad \forall \ (i,i') \in O \times O \label{eq8}
\end{align}
\end{figure*}

In this formulation, the objective is to minimize the makespan, as indicated in Equation \ref{eq1}. Constraint \ref{eq2} ensures that each operation is assigned to exactly one machine. Constraint \ref{eq3} enforces the precedence relationships between consecutive operations. Constraints \ref{eq4} and \ref{eq5} prevent overlapping of operations scheduled on the same machine. Constraint \ref{eq6} defines the makespan, while Constraints \ref{eq7} and \ref{eq8} specify the domains of the decision variables.

\begin{align}
& \min \ r_{max}& \label{eq9}\\
& c_{max} \leq C & \quad  \label{eq10}\\
& r_{max}\geq  \sum_{i \in O, \ s.t \ k \in R_i} p_{k,i} \cdot \alpha_{k,i} &  \quad \forall \ k \in R \label{eq11}
\end{align}

In the lexicographic approach, once the makespan has been minimized, the next objective is to minimize workload balance, as expressed in Equation \ref{eq9}. To implement this approach, the makespan variable must be constrained so that it does not exceed the previously obtained value (Constraint \ref{eq10}), and the maximum machine productive time must also be appropriately bounded (Constraint \ref{eq11}). The remaining constraints from the previous model remain unchanged.

\begin{align}
& r_{max} \leq RM & \quad   \label{eq12}
\end{align}

When workload balance is prioritized as the primary objective, the objective function in Equation \ref{eq1} is replaced with that in Equation \ref{eq9} in the proposed model. Additionally, the constraint defining the makespan (Constraint \ref{eq6}) is substituted by the one defining the maximum machine productive time (Constraint \ref{eq11}). In this case, during the second phase of the lexicographic approach, the objective becomes the minimization of the makespan (Equation \ref{eq1}). Instead of using Constraint \ref{eq10} to limit the makespan, Constraint \ref{eq12} is introduced to impose an upper bound on the allowed usage of each machine.

\subsubsection{Constraint Programming Approach}
\label{subsec:cp}

Constraint Programming (CP) is a paradigm for solving combinatorial optimization problems \cite{laborie2009ibm}, particularly effective in complex scheduling contexts such as the FJSSP \cite{da2019industrial}. CP allows problems to be modeled declaratively by specifying decision variables, their domains, and the constraints that generate feasible solutions. 

In this FJSS context, each job comprises a sequence of operations, where each operation may be assigned to a subset of eligible machines, each with distinct processing times. The problem includes two primary constraints (see algorithm \ref{alg:cp_fjssp}). On the one hand, precedence constraints enforce the prescribed order of operations within each job. On the other hand, resource constraints prevent overlapping operations on the same machine.

For benchmarking, we implemented the CP model using two state-of-the-art solvers: (i) Google OR-Tools CP-SAT \cite{cpsatlp}, a robust open-source constraint programming engine known for its efficiency in solving scheduling problems, and (ii) IBM CPLEX CP Optimizer (DoCplex) \cite{cplex2009v12}, a commercial optimization suite that offers advanced support for scheduling via interval and sequence variables. These state-of-the-art CP solvers internally apply advanced techniques—combining constraint propagation with heuristic-guided search strategies—to prune the search space and construct feasible solutions. 

In the proposed pseudocode \ref{alg:cp_fjssp}, the two sequential objectives are: (1) minimizing the makespan, denoted as $C_{\max}$, and (2) minimizing the maximum workload across machines, denoted as $L_{\max}$.

For each operation $o_{ji}$ of job $j \in J$, assigned to machine $m \in M_{ji}$, the model defines optional \textit{interval variables} $iv_{jim}$ with fixed durations $p_{jim}$. Each operation can be scheduled on a subset of eligible machines $M_{ji}$. This flexibility is captured via the \textit{alternative} constraint, which connects the operation to its possible machine-specific interval assignments. The scheduling order of operations within the same job is enforced using \textit{end before start} precedence constraints. A   \textit{no overlap} constraint is applied to all intervals scheduled on the same machine to prevent machine conflicts.

The total load on each machine $m \in M$ is computed using the sum of processing times of all operations scheduled on that machine. The makespan $C_{\max}$ is defined as the latest completion time among all operations, and $L_{\max}$ is the maximum load across machines.

\begin{algorithm}
\caption{Constraint Programming Model for FJSSP with Lexicographic Objectives}
\label{alg:cp_fjssp}
\begin{algorithmic}[1]
\Require Set of jobs $J$, operations $O$, machines $M$, processing times $p_{jim}$
\Ensure the Schedule minimizes makespan and workload balance

\State \textbf{Define} interval variables $iv_{jim}$ for all $o_{ji} \in O$ and $m \in M_{ji}$
\State \textbf{Define} optional intervals with durations $p_{jim}$
\State \textbf{Define} decision variables $start_{ji}, end_{ji}$ for each operation $o_{ji}$

\ForAll{$o_{ji} \in O$}
    \State \textbf{Add} \textsc{Alternative}($o_{ji}$, $\{iv_{jim} \mid m \in M_{ji}\}$)
\EndFor

\ForAll{$j \in J$}
    \For{$i = 1$ \textbf{to} $|J_j|-1$}
        \State \textbf{Add} precedence constraint: \textsc{EndBeforeStart}($o_{ji}, o_{j,i+1}$)
    \EndFor
\EndFor

\ForAll{$m \in M$}
    \State \textbf{Add} \textsc{NoOverlap} constraint on $iv_{jim}$ scheduled on machine $m$
    \State \textbf{Define} $load_m \gets \sum_{ji} \textsc{PresenceOf}(iv_{jim}) \cdot p_{jim}$
\EndFor

\State \textbf{Define} makespan $C_{\max} = \max_{ji} \textsc{EndOf}(o_{ji})$
\State \textbf{Define} max machine productive time $L_{\max} = \max_{m \in M} load_m$


\end{algorithmic}
\end{algorithm}

To implement lexicographic optimization, the solver first minimizes. Once the optimal value is found, a constraint is added to fix its value by setting an upper bound. This ensures that any subsequent solution maintains the optimal primary objective value. Then, the model proceeds to minimize the secondary objective.

Overall, CP provides a scalable and expressive framework for solving FJSS instances with high precision and is especially advantageous in instances where logical dependencies and disjunctive resource constraints dominate.

\subsubsection{Baseline dispatching rule}

The third classical algorithm evaluated in the benchmarking study aims to define a heuristic baseline, based on a well-established dispatching rule methodology. This approach addresses the FJSSP by making sequential scheduling decisions according to predefined priority rules. Rather than solving the entire problem globally, the heuristic proceeds iteratively, selecting which job to schedule next and assigning its operations to available machines following specific criteria.

The algorithm \ref{alg:dr_fjssp} shows the decision-making process that begins by establishing a priority order for the jobs (line 3). Initially, this sequence is determined randomly to introduce variability and exploratory behavior. However, a more refined approach might incorporate other priority rules to improve performance in objectives such as tardiness or makespan.

Once a job is selected according to the priority, its operations are scheduled sequentially (line 7), respecting the defined route of the job. For each operation, the heuristic evaluates all machines capable of executing it (line 9). The objective is to identify the earliest possible time slot on any of these eligible machines, accounting for the current machine schedules and the completion time of the preceding operation (parameter $tc$). This ensures temporal feasibility and maintains the required sequence.

For each eligible machine, the heuristic calculates the potential start and finish times for the operation. It then selects the machine and the corresponding time slot, yielding the earliest possible finish time for the current operation (line 10). This greedy strategy aims to minimize the overall completion time of the operations. This is achieved by iteratively updating the parameters $bt$ (best time), $bm$ (best machine), and $bd$ (best duration) upon the identification of a better result (lines 12, 13 and 14 respectively). After assigning an operation, the machine’s schedule is updated accordingly, and the completion time becomes a constraint for scheduling the subsequent operation of the same job (lines 17 and 18). This process continues until all operations of all jobs are scheduled.

\begin{algorithm}
\caption{Baseline dispatching rule}
\label{alg:dr_fjssp}
\begin{algorithmic}[1]
\State \textbf{Input} Set of jobs $J$, operations $O_j$, machines $M_o$, processing times $p_{om}$
\State \textbf{Output} Sol

\State $J_p \gets prioritization(J)$
\State $Sol \gets \emptyset$

\ForAll{$O_j \in J_p$}
    \State $tc \gets 0$
    \ForAll{$o \in O_j$}  
        \State $bt \gets \infty$
        \ForAll{$m \in M_o$}  
            \State $t \gets earliesFeasibletTime(m, p_{om}, tc, Sol)$
            \If{$t < bt$}
                \State $bt \gets t$
                \State $bm \gets m$
                \State $bd \gets p_{om}$
            \EndIf
        \EndFor
        \State $Sol \gets Sol \cup (bt, bm, bd)$
        \State $tc \gets bt + bd$
    \EndFor
\EndFor
\end{algorithmic}
\end{algorithm}

\subsubsection{Advances in AI techniques}
\label{subsec:AItechniques}

Recent advances in Artificial Intelligence, particularly in Deep Learning and RL, have introduced promising paradigms for solving combinatorial optimization problems such as the FJSS problem \cite{lei2022multi, song2022flexible}. These techniques leverage historical data to learn scheduling patterns, enabling models to generalize across problem instances and adapt to dynamic environments. Unlike traditional methods that exhaustively explore the solution space, these approaches learn approximate policies that reduce computational complexity, accelerating decision-making, often with competitive or superior results. RL has emerged as a promising framework for job assignment and scheduling problems. 

In this benchmark, we have selected two recent state-of-the-art RL approaches applied to the well-established Brandimarte dataset \cite{Brandimarte1993}. (1) Ho et al. \cite{ho2024residual} introduced a novel deep reinforcement learning method which improves efficiency by filtering out completed jobs and occupied machines, allowing the model to focus exclusively on the remaining decision space. The approach employs graph neural networks to represent the scheduling state and acts as a construction heuristic. (2) Echeverria et al. \cite{echeverria2025leveraging} proposed a hybrid learning approach that integrates CP with RL. In this work, a deep model is trained on optimal solutions generated via CP, reducing the need for extensive exploration. Additionally, the CP model is embedded within the RL framework to refine intermediate solutions. 

Given the strong empirical results reported by both methods on benchmark datasets, we include them as RL baselines in our comparative study. Since the original implementations were not publicly released, the performance values are directly extracted from the respective publications.

\section{Results and Discussion}
\label{sec: results}
This section presents a comprehensive analysis of the experimental results obtained in evaluating our proposed approach to the FJSSP. The analysis is divided into three key components: an ablation study, a benchmarking evaluation using a well-established dataset, and a robustness assessment involving non-linear setup times. In this sense, these components aim to assess the effectiveness and adaptability of our methodology under varying conditions.

Before presenting the main results, preliminary experiments were conducted to select the best-performing variant of our approach. While our methodology leverages the flexibility of
LLMs to enhance multi-objective optimization, the benchmarking results reported in this section are limited to the top-performing configuration. This selection process is fully
documented in B. 

All experiments were conducted on a personal computing setup consisting of an 8-core Intel i7 processor, 16 GB of RAM, and no dedicated GPU. This choice of hardware emphasizes the practicality and scalability of the approach, showing that high-quality results can be achieved even with limited computational resources. Furthermore, the models were executed via the services provided through their respective APIs, underlining the accessibility and ease of integration of these tools into standard computing environments.

\subsection{Ablation Study}
\label{subsec:AblationStudy}

This experimental technique, widely used in machine learning research, systematically removes specific components of an algorithm to evaluate their impact on overall performance. Comparing the proposed model with a variant that excludes the reflection mechanism can demonstrate the effectiveness in enhancing solution diversity and guiding search processes more efficiently. This experiment aims to support the novelty of our approach and highlight the relevance of incorporating self-reflective strategies into the framework.

\begin{figure*}[ht]
\centering
\subfloat[HyperVolume (HV)]{\includegraphics[width=0.4\textwidth]{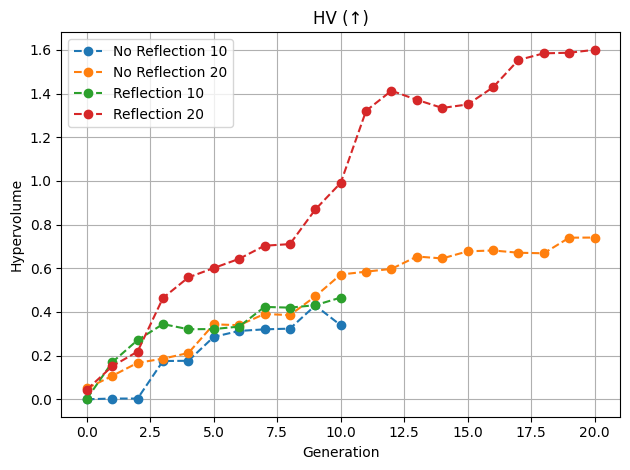}%
\label{fig:reflectioncomparison_HV}}
\hfil
\subfloat[Inverted Generational Distance (IGD)]{\includegraphics[width=0.4\textwidth]{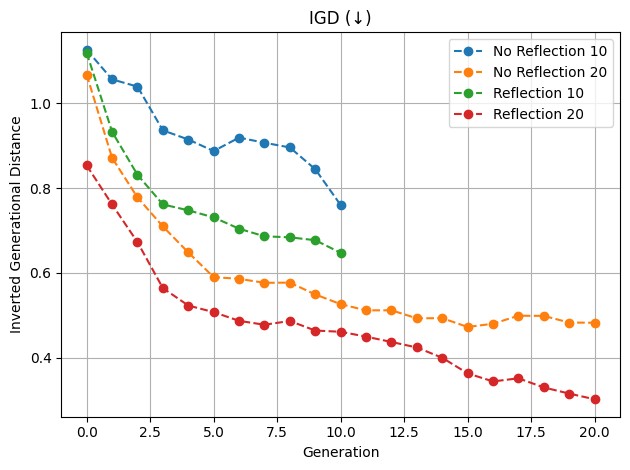}%
\label{fig:reflectioncomparison_IGD}}
\caption{Comparison between the reflective and non-reflective approaches using different population sizes and iterations.}
\label{fig: reflectioncomparison}
\end{figure*}

In this phase, two experimental configurations were defined to examine the consistency and scalability of the reflection mechanism under different conditions. The first configuration was initialized with a population size of 30 individuals, and subsequent generations were limited to a population size of 10, improved over 10 iterations. In the second configuration, the initial population was increased to 60 individuals, with a generation size of 20 and 20 iterations. All experiments were evaluated following the criteria mentioned in the subsection \ref{subsec: eval}.

The experiments were carried out by running the algorithm on the Brandimarte dataset. As previously mentioned, this experiment involves conflicting objectives: \textit{makespan} and \textit{workload balance}. Additionally, to test the capability of clustering reflection to handle challenging multi-objective problems, \textit{operation separation} was introduced as a third objective. This objective measures the time distance between operations of the same job, a typical goal in the FJSSP to enhance scheduling efficiency.


The figures \ref{fig: reflectioncomparison} present the HV and IGD values across generations for both algorithm variants in both experimental instances. Higher HV values indicate better convergence and diversity of solutions. In this sense, the evolution of HV in the figure \ref{fig:reflectioncomparison_HV} illustrates the improvement in solution diversity and convergence quality when using the reflection mechanism, as evidenced by consistently higher HV values in both instances. Conversely, lower IGD values in the figure \ref{fig:reflectioncomparison_IGD} indicate better convergence toward the Pareto front. The figure on the right shows that the reflection-enhanced approach consistently has lower IGD, demonstrating improved convergence behavior.

The observed results provide empirical evidence demonstrating that the reflection mechanism's inclusion significantly improves both convergence (HV) and proximity to the Pareto front (IGD). This component maintains diversity and prevents premature convergence by generating new individuals based on clustering and introspection of the search space. These findings support our hypothesis that integrating LLM-powered reflection enhances the robustness and adaptability of multi-objective optimization algorithms.

\subsection{Brandimarte Benchmark}
\label{subsec:brandimarte-benchmark}

In this section, we present the benchmarking results of our proposed methodology on the widely recognized Brandimarte dataset for the FJSSP. 

Given the multi-objective nature of our study, we adopted a lexicographic optimization approach for the mathematical methods. This leads to developing two configurations per model, one prioritizing makespan ($mb$), and another prioritizing workload balance ($bm$). As a result, six baseline models were used for benchmarking: $MILP_{mb}$, $MILP_{bm}$, $OR_{mb}$, $OR_{bm}$, $DC_{mb}$, and $DC_{bm}$, where OR and DC refer to constraint programming solvers using OR-Tools and DoCplex Optimizer, respectively. Due to the computational complexity of the FJSSP, time limits were set on the mathematical solvers. For the MILP model, we solved the problems using Gurobi Optimizer version 11.0.1 and allowed a maximum solving time of 30 minutes per objective, resulting in a total time of one hour per instance. In contrast, CP models are known for their efficiency in handling scheduling constraints, so a limit of 5 minutes was assigned per objective. Additionally, to provide a baseline for comparison, we executed a traditional greedy dispatching rule (DR) ten times for each instance, and from these runs, we selected the best-performing result for each objective.

Regarding the RL approaches, Ho et al. \cite{ho2024residual} ($RL_1$) and Echeverria et al. \cite{echeverria2025leveraging} ($RL_2$), these works only report results for the makespan minimization problem in instances from 1 to 10. Thus, both models are excluded from the multi-objective benchmarking analysis, although they are still referenced in the single-objective comparison.

Finally, following the experimentation performed in the LLM model selection (see Appendix B), 13 heuristics were obtained in the non-dominated front using Gemini 2.0 Flash in the REMoH framework. To perform the comparison of our approach, we selected the best result on the makespan-balance objective pair for each instance of Brandimarte among these 13 evolved heuristics. These results are reflected in the table \ref{tab: Benchmark_Makespan} and \ref{tab: Benchmark_Balance}.

For the evaluation of makespan and workload balance optimization, we report both the achieved objective values and the best-known lower bounds (LBs) per instance, selecting the highest LB value reported across all MILP and CP configurations, as well as studies in literature \cite{ho2024residual, echeverria2025leveraging, 10367980}.

\begin{table*}[ht]
\centering
\caption{Benchmarking of Makespan optimization on Brandimarte Instances, comparing mathematical models, Ho et al. \cite{ho2024residual} ($RL_1$), Echeverria et al.\cite{echeverria2025leveraging} ($RL_2$), and three heuristics obtained by REMoH} 
\label{tab: Benchmark_Makespan}
\begin{tabular}{lccccccc|c}
& $MILP_{mb}$ & $DC_{mb}$ & $OR_{mb}$ & $DR$ & $RL_1$ & $RL_2$ & REMoH & LB \\
\hline
Mk01 & \textbf{40} & \textbf{40} & \textbf{40} & 64 & 42 & \textbf{40} & 41 & 40 \\
Mk02 & \textbf{26} & \textbf{26} & \textbf{26} & 41 & 29 & 29 & 32 & 26 \\
Mk03 & \textbf{204} & \textbf{204} & \textbf{204} & 329 & \textbf{204} & \textbf{204} & \textbf{204} & 204 \\
Mk04 & 61 & \textbf{60} & \textbf{60} & 117 & 67 & 64 & 72 & 60 \\
Mk05 & 180 & 173 & \textbf{172} & 240 & 177 & 176 & 182 & 172 \\
Mk06 & 66 & 59 & \textbf{58} & 91 & 71 & 73  & 73 & 57 \\
Mk07 & 151 & \textbf{139} & \textbf{139} & 192 & 149 & 156 & 161 & 139 \\
Mk08 & \textbf{523} & \textbf{523} & \textbf{523} & 661 & \textbf{523} & \textbf{523} & \textbf{523} & 523 \\
Mk09 & 311 & \textbf{307} & \textbf{307} & 438 & 314 & \textbf{307} & 323 & 307 \\
Mk10 & 255 & \textbf{202} & 211 & 388 & 218 & 236 & 225 & 189 \\
Mk11 & 629 & 612 & \textbf{609} & 757 & - & - & 658 & 609 \\
Mk12 & \textbf{508} & \textbf{508} & \textbf{508} & 674 & - & - & 534 & 508 \\
Mk13 & 433 & 406 & \textbf{403} & 586 & - & - & 471 & 369 \\
Mk14 & 702 & \textbf{694} & \textbf{694} & 1072 & - & - & 707 & 694 \\
Mk15 & 360 & \textbf{333} & 344 & 552 & - & - & 422 & 333 \\
\hline
GAP & 6.46 & \textbf{1.43} & 1.73 & 54.78 & 8.05 & 8.57 & 12.60 & 
\end{tabular}
\end{table*}

\begin{table}[ht]
\centering
\caption{Benchmarking of Workload Balance  on Brandimarte Instances comparing mathematical models, and three heuristics obtained by REMoH}
\label{tab: Benchmark_Balance}
\begin{tabular}{lccccc|c}
& $MILP_{bm}$ & $DC_{bm}$ & $OR_{bm}$ & $DR$ & REMoH & LB \\
\hline
Mk01 & \textbf{36} & \textbf{36} & \textbf{36} & 55 & 37 & 36 \\
Mk02 & \textbf{26} & \textbf{26} & \textbf{26} & 31 & 31 & 26 \\
Mk03 & \textbf{204} & \textbf{204} & \textbf{204} & 274 & \textbf{204} & 204 \\
Mk04 & \textbf{60} & \textbf{60} & \textbf{60} & 103 & 67 & 60 \\
Mk05 & \textbf{172} & \textbf{172} & \textbf{172} & 194 & 180 & 172 \\
Mk06 & \textbf{48} & \textbf{48} & \textbf{48} & 70 & 68 & 48 \\
Mk07 & \textbf{139} & \textbf{139} & \textbf{139} & 158 & 161 & 139 \\
Mk08 & \textbf{523} & \textbf{523} & \textbf{523} & 542 & \textbf{523} & 523 \\
Mk09 & \textbf{299} & \textbf{299} & \textbf{299} & 347 & 307 & 299 \\
Mk10 & \textbf{189} & \textbf{189} & \textbf{189} & 282 & 223 & 188 \\
Mk11 & \textbf{609} & \textbf{609} & \textbf{609} & 652 & 658 & 609 \\
Mk12 & \textbf{508} & \textbf{508} & \textbf{508} & 572 & 534 & 508 \\
Mk13 & \textbf{382} & \textbf{382} & \textbf{382} & 477 & 470 & 382 \\
Mk14 & \textbf{694} & \textbf{694} & \textbf{694} & 903 & 707 & 694 \\
Mk15 & \textbf{332} & \textbf{332} & \textbf{332} & 406 & 422 & 332 \\
\hline
GAP & \textbf{0.04} & \textbf{0.04} & \textbf{0.04} & 27.79 & 12.15 & 

\end{tabular}
\end{table}

The benchmarking results in Tables \ref{tab: Benchmark_Makespan} and \ref{tab: Benchmark_Balance} demonstrate the performance of our proposed REMoH approach relative to both classical optimization models and state-of-the-art RL methods on the Brandimarte dataset. The GAP values on the tables show the mean GAP values of each model over all instances, where GAP stands for the relative percentual difference between the objective value and the lower-bound.

In this sense, REMoH achieves competitive performance in minimizing makespan ($12.60$\% of GAP), obtaining results close to those of the RL-based models \cite{ho2024residual, echeverria2025leveraging}, while significantly outperforming the traditional dispatching rule baseline in all instances. It is important to note that RL approaches are designed to optimize only the makespan, while REMoH considers both makespan and workload balance during the search process. Additionally, RL models could not obtain results for instances larger than Mk10.

Results show that mathematical models achieve strong performance in their respective prioritized objectives due to the lexicographic optimization, for instance, $DC_{mb}$ obtained a $1.43$\% of mean GAP. In contrast, REMoH is designed to optimize both objectives, obtaining a more holistic view of solution quality. This difference explains why certain objective-specific methods may occasionally outperform REMoH in their dominant metric, e.g., $MILP_{mb}$ in makespan or $MILP_{bm}$ in workload balance. In this sense, heuristics obtained by REMoH provide consistently balanced trade-offs across both dimensions.

\begin{figure*}[ht]
\centering
\subfloat[Mk6 instance: 10 jobs, 10 machines.]{\includegraphics[width=0.5\textwidth]{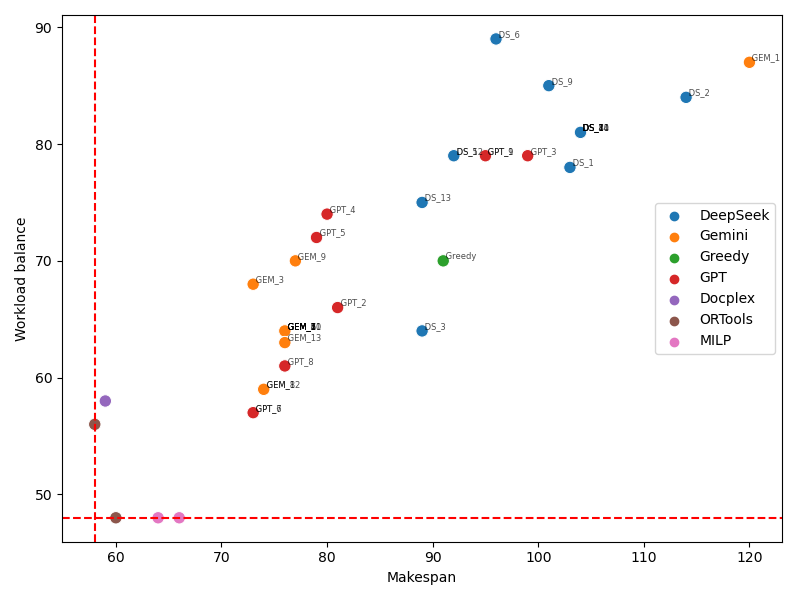}%
\label{fig:first}}
\hfil
\subfloat[Mk10 instance: 20 jobs, 15 machines.]{\includegraphics[width=0.5\textwidth]{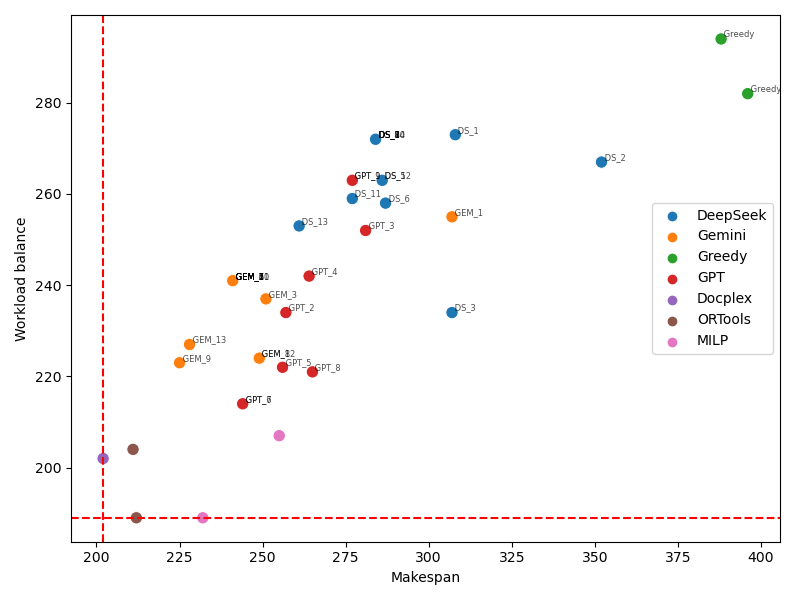}%
\label{fig:second}}
\caption{Illustrative results on Brandimarte instances. The table compares outcomes from MILP, DoCplex, OR-Tools, and heuristics derived from our LLM-integrated model, which incorporates DeepSeek (DS), Gemini 2.0 Flash (GEM), and GPT-4o. (GPT)}
\label{fig:figures}
\end{figure*}

\begin{figure}[ht]
  \centering
  \includegraphics[width=0.45\textwidth]{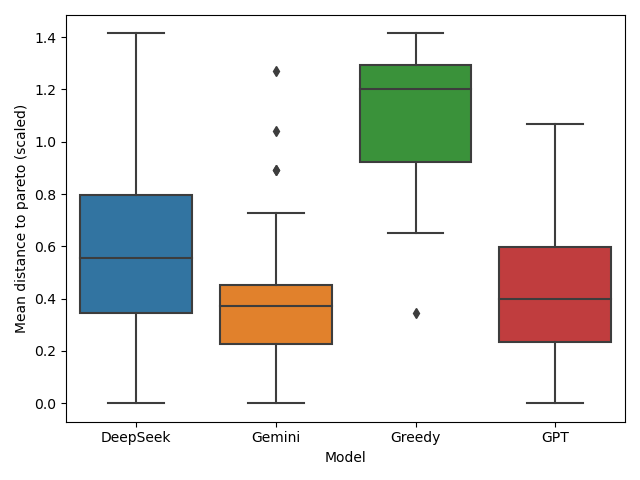}
  \caption{Boxplot of the mean Euclidean distances to best-known solutions for different approaches.}
  \label{fig:bpmodels}
\end{figure}

The mean GAP value is a representative performance metric, nevertheless, we aimed to analyze the benchmarking process further. Therefore, we measured the distance of each heuristic-generated solution to the best-known reference solutions, typically obtained from mathematical models. To this end, the objective values for each instance were individually normalized, and the Euclidean distance from each heuristic-generated point to its corresponding reference point was calculated (see Figure \ref{fig:bpmodels}). The average normalized distance was then used as a comparative metric across the different heuristics, including the LLM-generated methods described in Appendix B, and the greedy DR.

The model with the lowest average distance to the Pareto front was Gemini, followed by GPT, DeepSeek, and finally the greedy DR. These results suggest that heuristics generated by REMoH using Gemini are more consistent across unseen problem instances. Furthermore, GPT consistently outperformed DeepSeek in generalization to new data, despite DeepSeek achieving better results at higher generations during the training phase (see Appendix B).

To further contextualize our results, we compared the average computational times for each method. Among the exact solvers, MILP averaged 1661.5s, CP with DoCplex 230.9s, and CP with OR-Tools 148.4s. These optimization times show the lexicographic optimization settings and the time limits imposed per objective. Notably, CP approaches consistently achieved lower optimality GAPs in significantly shorter runtimes compared to MILP.

As expected, the greedy DR was among the fastest due to its simplicity, with an average runtime of 0.002s. Interestingly, in the REMoH framework, GPT-generated heuristics ran even faster (0.001s), followed by DeepSeek (0.012s) and Gemini (0.167s). While Gemini showed the best solution quality, this came at the cost of increased computational complexity, reflected in longer execution times.

According to recent findings in the literature \cite{DAUZEREPERES2024409}, CP-based methods have outperformed MILP in solving the FJSSP. This trend is reflected in our experiments, where solutions from CP often dominate those obtained using MILP for both objectives. However, MILP was occasionally more time-efficient when workload balance was prioritized over makespan, achieving comparable solutions in less time.

To illustrate these dynamics, Figure \ref{fig:figures} presents a plot with both objectives for two representative instances, Mk06 and Mk10. These results reveal that for smaller instances, the greedy DR remains competitive and can even outperform some learned heuristics. However, as instance complexity increases, the REMoH-generated heuristics become more relevant. Conversely, MILP loses dominance as instance size grows, and CP approaches tend to produce superior solutions.

In conclusion, REMoH obtains stable performance across all instances, often matching or approaching the best-known lower bounds derived from exact methods. These findings validate the effectiveness of our LLM-driven heuristic evolution framework as a robust alternative to classical and learning-based scheduling methods, but also capable of addressing complex multi-objective problems.

\subsection{Flexibility Evaluation}
\label{subsec:robustness-evaluation}

This section aims to demonstrate the flexibility of the proposed solution. To this end, the baseline FJSS formulation is extended by incorporating an additional constraint that simulates a more complex and realistic scheduling environment.

Therefore, we introduce a well-established generalization named \textit{Sequence-Dependent Setup Times} (SDST) \cite{mousakhani2013sequence}. In practice, setup time refers to the duration required to configure a resource before it can execute a specific operation \cite{caldeira2021simheuristic}. In SDST scenarios, this time usually depends on both the incoming operation and the operation previously completed on the same machine. In this case, to enhance the realism of this extension, we embed a nonlinear behavior in the setup time by modeling it as a function of machine idle time. This reflects practical phenomena observed in manufacturing systems where prolonged machine inactivity can increase startup costs, for instance, due to temperature sensitivity. Specifically, we define a setup time function $S(t)$ as follows:

\[
S(t) =
\begin{cases}
0 & \text{if } t < 1 \\
10 \cdot (1 - e^{-0.5(t - 1)}) & \text{if } 1 \leq t < 10 \\
20 & \text{if } t \geq 10
\end{cases}
\]

\begin{figure*}[ht]
\centering
\subfloat[HyperVolume (HV)]{\includegraphics[width=0.5\textwidth]{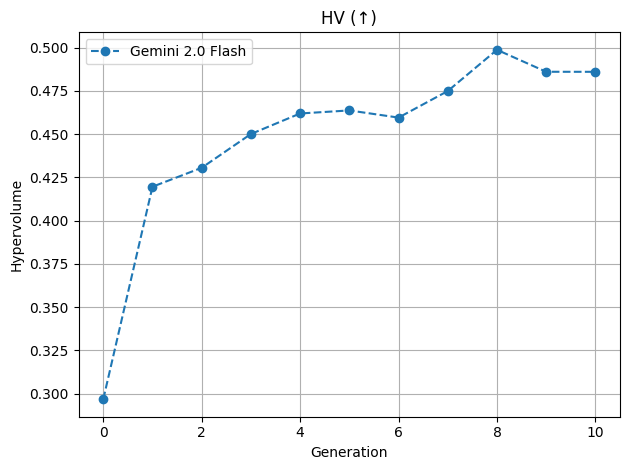}%
\label{fig:sdst-hv}}
\hfil
\subfloat[Inverted Generational Distance (IGD)]{\includegraphics[width=0.5\textwidth]{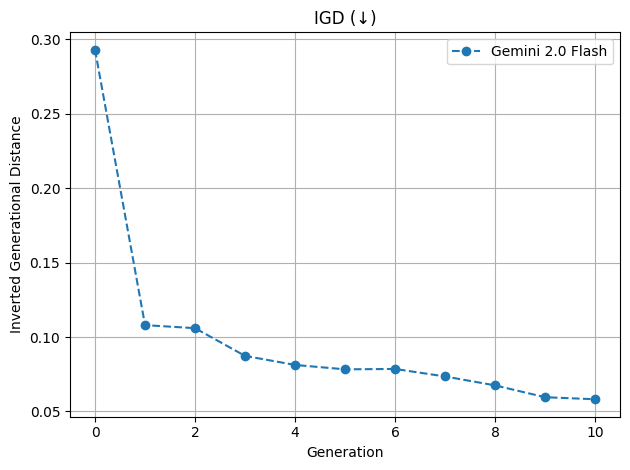}%
 \label{fig:sdst-igd}}
\caption{Heuristic evolution performance validation with non-linear Sequence-Dependent Setup Times (SDST).}
\label{fig: sdstcomparison}
\end{figure*}

The integration of this nonlinear constraint poses substantial challenges for traditional optimization paradigms. MILP frameworks, for instance, cannot directly model nonlinear expressions without resorting to approximations that affect solution quality. Although constraint programming provides flexibility, incorporating these nonlinearities requires specialized propagators and search heuristics, which makes implementation more complex.

In this sense, more advanced techniques such as RL are also affected. Introducing a novel constraint requires creating a new training dataset, redefining the reward function, and fully retraining the policy network. This makes the approach computationally expensive and requires domain expertise to avoid convergence issues or suboptimal generalization.

By contrast, the methodology presented in this study is inherently capable of integrating nonlinear and context-sensitive constraints without requiring extensive model restructuring. This intrinsic flexibility underscores the robustness of our approach and highlights its potential for deployment in complex and dynamic production environments. To empirically validate this claim, we conducted an experiment incorporating the proposed nonlinear setup time function into our scheduling model.

To achieve this, minimal adjustments were made, slightly modifying the prompts given to the LLM by adding the constraint explanation and the method to calculate the delay. Finally, the feasibility checking function was adapted to include this new condition.


The initial population size was set to 30, with 10 individuals used for subsequent generations and iterations. The model was executed on the Brandimarte dataset, which, as previously discussed, is well-suited for evaluating heuristic performance across problems of varying sizes. To measure the performance, the aforementioned evaluation process was followed (see Section \ref{subsec: eval}).

As shown in figures \ref{fig:sdst-hv} and \ref{fig:sdst-igd}, REMoH can adapt to dynamic constraints, which makes it highly flexible and applicable for real-world optimization problems, where complex constraints are often challenging to handle with traditional heuristics and RL.

\section{Conclusions}
\label{sec:Conclusions}

This research paper presents a novel framework named  \textit{Reflective Evolution of Multi-objective Heuristics} (REMoH) that incorporates LLMs into the heuristic design process. The proposed methodology aims to leverage the capacity of generalization, abstraction, and reasoning capabilities  to evolve domain-agnostic, interpretable heuristics. In this sense, this approach facilitates efficient solution space exploration through adaptive evolution.

The main innovative component of the framework is the introduction of a \textit{Reflection Mechanism}, detailed in Section \ref{sec: method}, which enables long-term reflection in the evolutionary process. This mechanism performs clustering in the objective space to analyze population structure and uses introspective LLM reasoning to generate adapted heuristics that improve exploration and mitigate premature convergence. The effectiveness of this component is empirically validated in the ablation study (see Section \ref{subsec:AblationStudy}), demonstrating notable improvements in convergence, diversity, and proximity to the Pareto front.

The optimization capability of the approach was further evaluated in a benchmarking study conducted on the Brandimarte dataset. First, a model selection phase was performed to identify the most suitable LLM, then the best-performing heuristics were obtained after the evolving process on Barnes and Dauzere instances. These heuristics were compared against classical mathematical programming approaches (MILP, CP), a baseline heuristic based on dispatching rules, and state-of-the-art RL methods. The results indicate that our proposed framework can generate heuristics that achieve competitive or superior performance in terms of makespan, outperforming RL-based approaches in some Brandimarte instances (see Table \ref{tab: Benchmark_Makespan}). Additionally, the algorithm proposes solutions with significantly lower computational overhead. While mathematical programming-based models obtained optimal solutions for small instances, they require significant runtime costs, making them less practical for real-time or large-scale scenarios. Unlike exact methods, which often suffer from scalability issues and require substantial expertise in both problem modeling and optimization techniques, the proposed approach relies solely on prompt engineering. This significantly lowers the entry barrier, as it does not require prior knowledge of the underlying problem or the employed technique.

Finally, a generalization of the FJSSP was formulated to assess the flexibility and robustness. This variant incorporates non-linear SDST as a nonlinear function related to machine idle time. Traditional mathematical models typically struggle with integrating such constraints without extensive reformulation. In contrast, this methodology can adapt to this problem variant, successfully evolving heuristics capable of solving the generalized FJSSP efficiently (see Section \ref{subsec:robustness-evaluation}). This demonstrates the potential for handling complex, dynamic, and constraint-rich optimization environments.

In conclusion, this work highlights the potential of LLMs in advancing multi-objective optimization, particularly in enabling adaptive and interpretable heuristic design. The proposed REMoH framework offers a flexible alternative to conventional methods while achieving competitive performance with less modeling effort. Future research may explore the evolution of full metaheuristic architectures, extending to high-level algorithm generation.

\section*{Copyright}
© 2025 IEEE. Personal use of this material is permitted. Permission from IEEE must be obtained for all other uses, in any current or future media, including reprinting/republishing this material for advertising or promotional purposes, creating new collective works, for resale or redistribution to servers or lists, or reuse of any copyrighted component of this work in other works.

\appendices

\section{Pseudocode of the Proposed Multi-Objective EoH Algorithm}
\label{appendix: pseudocode_A}

This appendix provides the detailed pseudocode of the proposed Multi EoH algorithm. This hybrid framework integrates NSGA-based multi-objective evolutionary optimization with LLMs for heuristic generation. The pseudocode \ref{alg:MEoH} encapsulates the entire optimization pipeline. This detailed algorithmic structure provides a transparent foundation for reproducibility and further research.

For each iteration $t \in {1, \dots, \mathcal{T}}$, the algorithm evolves a population of heuristics through LLM-guided generation, reflection, and selection mechanisms. The process begins with an initial population $\mathcal{P}_0$, either provided or generated from scratch using the LLM $\mathcal{L}$.

At each generation step, a subset of individuals is selected to form the parent population $P_{parents}$, which is then clustered into subgroups. These clusters serve as the basis for generating short reflections via $\texttt{Reflection}(\mathcal{C}, \mathcal{L})$, summarizing the strategic patterns observed. A more comprehensive reflection $\mathcal{R}_t$ is subsequently constructed by combining the short reflections with those from previous iterations.

Two types of offspring are generated:

Crossover-based heuristics, created by combining two selected parents $(p_1, p_2)$ guided by $\mathcal{R}_t$ and the LLM, account for $r_c \cdot \mathcal{P}$ new individuals.

Mutation-based heuristics, derived from a single elite parent $p_{best}$ using $\mathcal{R}_t$ and the LLM, account for $r_m \cdot \mathcal{P}$ individuals.

The new population $P_t$ is formed by merging the parent and offspring populations. All generated individuals across iterations are collected into a global archive $P^*$.

At the end of the evolutionary process, the global Pareto front $\mathcal{F}$ is extracted from $P^*$, representing the most effective heuristics found in terms of multiple objectives.

\begin{algorithm}
\caption{Proposed Multi-Evolution of Heuristics}
\label{alg:MEoH}
\begin{algorithmic}[1]
\State \textbf{Input:} Initial population size $\mathcal{I}$, Population size $\mathcal{P}$, Iterations $\mathcal{T}$, Initial population $\mathcal{P_0}$, Pre-trained LLM $\mathcal{L}$, Crossover rate $r_c$, Mutation rate $r_m$
\State \textbf{Output:} Heuristics in global Pareto front $\mathcal{F}$

\If{$\mathcal{P}_0 = \emptyset$}
    \For{$i = 1, ..., \mathcal{I}$}
        \State $o \gets Generation(\mathcal{L})$
        \State $P_0 \gets P_0 \cup o$
    \EndFor
\EndIf
\State $P^* \gets P_0$
\For{$t = 1, ..., \mathcal{T}$}
    \State $P_{parents} \gets Selection(P_{t-1})$
    \State $\mathcal{C} \gets Clusterize(P_{parents})$
        
    \State $\mathcal{R}_{short} \gets Reflection(\mathcal{C}, \mathcal{L})$
    
    \State $\mathcal{R}_t \gets Reflection(\mathcal{C}, \mathcal{R}_{short}, \mathcal{R}_{t-1}, \mathcal{L})$
    \State $P_{sons} \gets \emptyset$
    \For{$j = 1, ..., r_c * \mathcal{P}$}
        \State $p_1, p_2 \gets ParentSelection(P_{parents})$
        \State $o \gets Crossover(p_1, p_2, \mathcal{R}_t, \mathcal{L})$
        \State $P_{sons} \gets P_{sons} \ \cup \ o$
    \EndFor
        
    \For{$j = 1, ..., r_m * \mathcal{P}$}
        \State $p_{best} \gets bestHeuristic(P_{parents})$
        \State $o \gets Mutation(p_{best}, \mathcal{R}_t, \mathcal{L})$
        \State $P_{sons} \gets P_{sons} \ \cup \ o$
    \EndFor

    \State $P_{t} \gets P_{parents} \ \cup \ P_{sons}$
    \State $P^* \gets P^* \ \cup \ P_{sons}$
\EndFor

\State $\mathcal{F} \gets GlobalParetoFront(P^*)$ 
\end{algorithmic}
\end{algorithm}

\section{LLM Selection}\label{apendix:llm_selection}

This appendix provides a comprehensive overview of the comparative evaluation conducted to select the most effective LLM in our Reflective Evolution of Multi-objective Heuristics (REMoH) framework. 

This study aimed to identify the LLM that consistently produces high-quality heuristic operators and reflections, thereby maximizing the performance of the proposed multi-objective scheduling methodology. The tested LLMs include \texttt{GPT-4o} \cite{openai2023gpt4}, \texttt{Gemini 2.0 Flash} \cite{gemini20flash}, and \texttt{DeepSeek-V3} \cite{deepseek}. These models were selected due to their competitive performance and accessibility in practical optimization workflows.

The comparison was performed using training on two independent datasets from the FJSSP benchmark literature, Dauzere and Barnes, and validating in the Brandimarte dataset. The experiments focused on determining which LLM most effectively supports the generation of high-quality Pareto fronts across these diverse conditions. The evaluation and selection criteria for this experimentation is explained in the subsection \ref{subsec: eval}. In this sense, the evaluation is performed in the same environment, with identical configurations across multiple runs. 

The optimization process was initialized with a population size of 60 individuals. Then, throughout the evolutionary process, a smaller working population of 20 individuals was maintained for each generation. Each run consisted of 20 iterations of the algorithm. During the experiments, heuristics were evaluated on a maximum time of 100 seconds. Regarding the evolutionary operators, crossover and mutation rates were fixed at 0.9 and 0.1, respectively.

\begin{figure}[ht]
  \centering
  \includegraphics[width=0.8\linewidth]{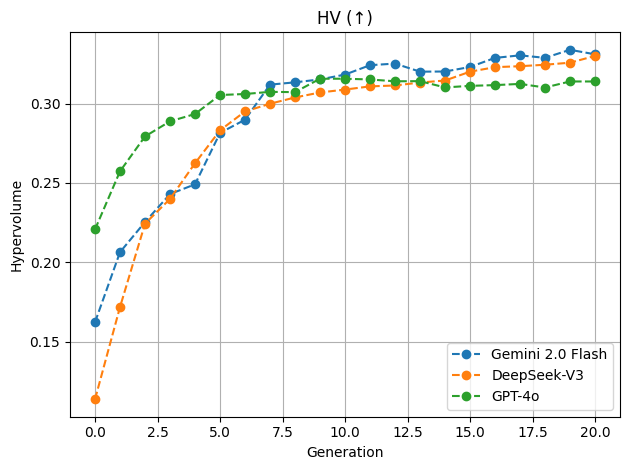}
  \caption{Comparison between different LLMs in Hypervolume (HV).}
  \label{fig:modelsHV}
\end{figure}

\begin{figure}[ht]
  \centering
  \includegraphics[width=0.8\linewidth]{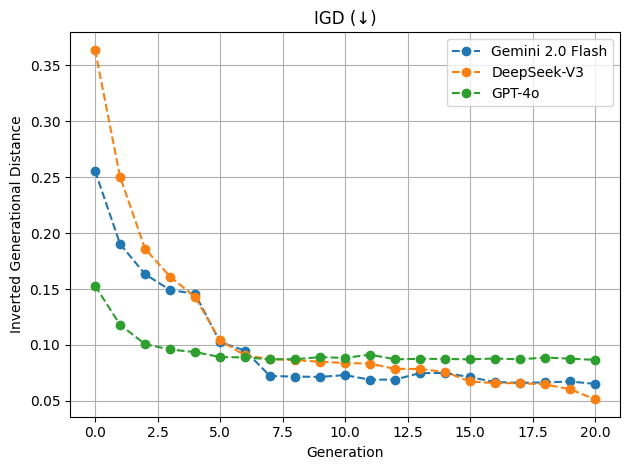}
  \caption{Comparison between different LLMs in Inverted Generational Distance (IGD).}
  \label{fig:modelsIGD}
\end{figure}

As shown in Figures \ref{fig:modelsHV} and \ref{fig:modelsIGD}, all models exhibit a significant improvement during the initial generations. After this early boost, \texttt{DeepSeek-V3} and \texttt{Gemini 2.0 Flash} follow a similar progression, gradually improving their performance over successive generations. In contrast, \texttt{GPT-4o} achieves strong results in the early stages but converges prematurely, making it increasingly difficult to improve in later iterations. As a result, although the final differences are relatively small, \texttt{GPT-4o} ends up with the lowest performance among the three models.

The Pareto front heuristics generated by each model have been evaluated on the Brandimarte datasets, with results for all instances presented in Tables \ref{tab:deepseek_results}, \ref{tab:gemini_results} and \ref{tab:gpt_results}. Based on this evaluation, \texttt{Gemini 2.0 Flash} was selected as the reference model for comparison against established benchmarking methods, due to its superior performance. The results of this comparison are presented in Section \ref{subsec:brandimarte-benchmark}.

\begin{table*}[ht]
\centering
\caption{Results of the Pareto front heuristics generated by REMoH framework on the Brandimarte Instances, via DeepSeek-V3 \cite{deepseek}.}
\label{tab:deepseek_results}
\resizebox{\textwidth}{!}{%
\begin{tabular}{llccccccccccccccccc}
 &   & mk01 & mk02 & mk03 & mk04 & mk05 & mk06 & mk07 & mk08 & mk09 & mk10 & mk11 & mk12 & mk13 & mk14 & mk15 \\ \hline
\multirow{3}{*}{1} & Makespan & 52 & 41 & 236 & 83 & 194 & 103 & 264 & 575 & 411 & 308 & 735 & 612 & 577 & 808 & 480 \\
& Balance & 43 & 37 & 216 & 67 & 194 & 78 & 234 & 523 & 312 & 273 & 684 & 563 & 577 & 781 & 397 \\
& Time (s.) & 0.0010 & 0.0020 & 0.0050 & 0.0020 & 0.0040 & 0.0048 & 0.0041 & 0.0071 & 0.0083 & 0.0091 & 0.0102 & 0.0099 & 0.0124 & 0.0135 & 0.0142 \\
\hline
\multirow{3}{*}{2} & Makespan & 59 & 48 & 251 & 107 & 208 & 114 & 292 & 605 & 421 & 352 & 773 & 650 & 672 & 996 & 480 \\
& Balance & 49 & 40 & 216 & 83 & 196 & 84 & 255 & 523 & 327 & 267 & 730 & 565 & 652 & 801 & 445 \\
& Time (s.) & 0.0015 & 0.0020 & 0.0057 & 0.0039 & 0.0050 & 0.0040 & 0.0040 & 0.0075 & 0.0088 & 0.0115 & 0.0080 & 0.0095 & 0.0136 & 0.0123 & 0.0189 \\
\hline
\multirow{3}{*}{3} & Makespan & 45 & 40 & 236 & 75 & 223 & 89 & 200 & 629 & 423 & 307 & 677 & 613 & 529 & 828 & 480 \\
& Balance & 43 & 33 & 204 & 66 & 178 & 64 & 191 & 533 & 307 & 234 & 644 & 524 & 499 & 746 & 354 \\
& Time (s.) & 0.0031 & 0.0026 & 0.0155 & 0.0061 & 0.0091 & 0.0115 & 0.0085 & 0.0344 & 0.0374 & 0.0468 & 0.0338 & 0.0411 & 0.0521 & 0.0855 & 0.0836 \\
\hline
\multirow{3}{*}{4} & Makespan & 48 & 40 & 227 & 73 & 189 & 104 & 215 & 523 & 337 & 284 & 663 & 537 & 489 & 694 & 468 \\
& Balance & 43 & 39 & 213 & 68 & 189 & 81 & 215 & 523 & 322 & 272 & 663 & 537 & 489 & 694 & 468 \\
& Time (s.) & 0.0010 & 0.0020 & 0.0057 & 0.0030 & 0.0040 & 0.0056 & 0.0048 & 0.0099 & 0.0109 & 0.0158 & 0.0117 & 0.0130 & 0.0238 & 0.0150 & 0.0230 \\
\hline
\multirow{3}{*}{5} & Makespan & 48 & 38 & 230 & 69 & 192 & 92 & 200 & 523 & 327 & 286 & 681 & 550 & 487 & 694 & 428 \\
& Balance & 42 & 37 & 227 & 68 & 192 & 79 & 200 & 523 & 307 & 263 & 681 & 550 & 487 & 694 & 428 \\
& Time (s.) & 0.0009 & 0.0020 & 0.0050 & 0.0020 & 0.0055 & 0.0050 & 0.0045 & 0.0092 & 0.0099 & 0.0136 & 0.0109 & 0.0141 & 0.0216 & 0.0195 & 0.0223 \\
\hline
\multirow{3}{*}{6} & Makespan & 48 & 38 & 241 & 69 & 192 & 96 & 200 & 523 & 322 & 287 & 681 & 550 & 487 & 694 & 428 \\
& Balance & 42 & 37 & 204 & 68 & 192 & 89 & 200 & 523 & 307 & 258 & 681 & 550 & 487 & 694 & 428 \\
& Time (s.) & 0.0021 & 0.0025 & 0.0060 & 0.0039 & 0.0040 & 0.0051 & 0.0052 & 0.0091 & 0.0132 & 0.0151 & 0.0122 & 0.0139 & 0.0207 & 0.0221 & 0.0246 \\
\hline
\multirow{3}{*}{7} & Makespan & 48 & 40 & 227 & 73 & 189 & 104 & 215 & 523 & 337 & 284 & 663 & 537 & 489 & 694 & 468 \\
& Balance & 43 & 39 & 213 & 68 & 189 & 81 & 215 & 523 & 322 & 272 & 663 & 537 & 489 & 694 & 468 \\
& Time (s.) & 0.0020 & 0.0017 & 0.0085 & 0.0031 & 0.0040 & 0.0051 & 0.0057 & 0.0101 & 0.0115 & 0.0155 & 0.0137 & 0.0115 & 0.0248 & 0.0197 & 0.0252 \\
\hline
\multirow{3}{*}{8} & Makespan & 48 & 40 & 227 & 73 & 189 & 104 & 215 & 523 & 337 & 284 & 663 & 537 & 489 & 694 & 468 \\
& Balance & 43 & 39 & 213 & 68 & 189 & 81 & 215 & 523 & 322 & 272 & 663 & 537 & 489 & 694 & 468 \\
& Time (s.) & 0.0010 & 0.0020 & 0.0076 & 0.0042 & 0.0047 & 0.0045 & 0.0056 & 0.0080 & 0.0138 & 0.0142 & 0.0129 & 0.0110 & 0.0250 & 0.0376 & 0.0236 \\
\hline
\multirow{3}{*}{9} & Makespan & 49 & 37 & 223 & 73 & 189 & 101 & 227 & 523 & 328 & 284 & 663 & 537 & 506 & 694 & 448 \\
& Balance & 43 & 37 & 213 & 68 & 189 & 85 & 227 & 523 & 307 & 272 & 663 & 537 & 506 & 694 & 448 \\
& Time (s.) & 0.0010 & 0.0041 & 0.0084 & 0.0040 & 0.0050 & 0.0070 & 0.0070 & 0.0110 & 0.0143 & 0.0194 & 0.0130 & 0.0113 & 0.0311 & 0.0308 & 0.0256 \\
\hline
\multirow{3}{*}{10} & Makespan & 48 & 40 & 227 & 73 & 189 & 104 & 215 & 523 & 337 & 284 & 663 & 537 & 489 & 694 & 468 \\
& Balance & 43 & 39 & 213 & 68 & 189 & 81 & 215 & 523 & 322 & 272 & 663 & 537 & 489 & 694 & 468 \\
& Time (s.) & 0.0017 & 0.0035 & 0.0080 & 0.0042 & 0.0040 & 0.0041 & 0.0051 & 0.0103 & 0.0111 & 0.0181 & 0.0118 & 0.0105 & 0.0229 & 0.0271 & 0.0200 \\
\hline
\multirow{3}{*}{11} & Makespan & 50 & 39 & 238 & 79 & 187 & 104 & 219 & 525 & 353 & 277 & 682 & 538 & 506 & 694 & 446 \\
& Balance & 43 & 39 & 223 & 68 & 185 & 81 & 214 & 523 & 307 & 259 & 681 & 537 & 506 & 694 & 428 \\
& Time (s.) & 0.0010 & 0.0040 & 0.0077 & 0.0043 & 0.0040 & 0.0050 & 0.0056 & 0.0090 & 0.0115 & 0.0213 & 0.0137 & 0.0110 & 0.0260 & 0.0303 & 0.0216 \\
\hline
\multirow{3}{*}{12} & Makespan & 48 & 38 & 230 & 69 & 192 & 92 & 200 & 523 & 327 & 286 & 681 & 550 & 487 & 694 & 428 \\
& Balance & 42 & 37 & 227 & 68 & 192 & 79 & 200 & 523 & 307 & 263 & 681 & 550 & 487 & 694 & 428 \\
& Time (s.) & 0.0030 & 0.0041 & 0.0074 & 0.0031 & 0.0030 & 0.0054 & 0.0050 & 0.0081 & 0.0109 & 0.0185 & 0.0125 & 0.0122 & 0.0267 & 0.0590 & 0.0202 \\
\hline
\multirow{3}{*}{13} & Makespan & 47 & 40 & 232 & 69 & 186 & 89 & 215 & 523 & 341 & 261 & 663 & 537 & 489 & 694 & 448 \\
& Balance & 43 & 39 & 204 & 68 & 185 & 75 & 215 & 523 & 307 & 253 & 663 & 537 & 489 & 694 & 448 \\
& Time (s.) & 0.0009 & 0.0040 & 0.0070 & 0.0030 & 0.0033 & 0.0055 & 0.0043 & 0.0077 & 0.0098 & 0.0156 & 0.0100 & 0.0149 & 0.0262 & 0.0286 & 0.0183 \\
\hline
\multirow{3}{*}{14} & Makespan & 48 & 40 & 227 & 73 & 189 & 104 & 215 & 523 & 337 & 284 & 663 & 537 & 489 & 694 & 468 \\
& Balance & 43 & 39 & 213 & 68 & 189 & 81 & 215 & 523 & 322 & 272 & 663 & 537 & 489 & 694 & 468 \\
& Time (s.) & 0.0009 & 0.0031 & 0.0080 & 0.0035 & 0.0040 & 0.0049 & 0.0055 & 0.0071 & 0.0106 & 0.0173 & 0.0118 & 0.0111 & 0.0244 & 0.0273 & 0.0238 \\
\hline

\end{tabular}
}
\end{table*}

\begin{table*}[ht]
\centering
\caption{ Results of the Pareto front heuristics generated by REMoH framework on the Brandimarte Instances, via Gemini 2.0 Flash \cite{gemini20flash}.}
\label{tab:gemini_results}
\resizebox{\textwidth}{!}{%
\begin{tabular}{llccccccccccccccccc}
 &   &  mk01 & mk02 & mk03 & mk04 & mk05 & mk06 & mk07 & mk08 & mk09 & mk10 & mk11 & mk12 & mk13 & mk14 & mk15 \\ \hline
\multirow{3}{*}{1} & Makespan & 53 & 39 & 229 & 77 & 187 & 120 & 206 & 532 & 374 & 307 & 683 & 548 & 534 & 769 & 436 \\
& Balance & 40 & 38 & 216 & 67 & 181 & 87 & 206 & 523 & 299 & 255 & 683 & 548 & 534 & 734 & 398 \\
& Time (s.) & 0.0012 & 0.0030 & 0.0106 & 0.0040 & 0.0040 & 0.0076 & 0.0070 & 0.0110 & 0.0164 & 0.0367 & 0.0154 & 0.0147 & 0.0304 & 0.0310 & 0.0424 \\
\hline
\multirow{3}{*}{2} & Makespan & 41 & 32 & 221 & 74 & 186 & 76 & 161 & 523 & 323 & 241 & 683 & 534 & 471 & 707 & 422 \\
& Balance & 37 & 31 & 220 & 74 & 181 & 64 & 161 & 523 & 307 & 241 & 683 & 534 & 470 & 707 & 422 \\
& Time (s.) & 0.0102 & 0.0231 & 0.0785 & 0.0382 & 0.0417 & 0.1411 & 0.0690 & 0.1449 & 0.2377 & 0.4556 & 0.1982 & 0.1937 & 0.5520 & 0.4217 & 0.6594 \\
\hline
\multirow{3}{*}{3} & Makespan & 46 & 33 & 204 & 72 & 182 & 73 & 164 & 527 & 351 & 251 & 658 & 558 & 506 & 752 & 440 \\
& Balance & 42 & 33 & 204 & 67 & 180 & 68 & 164 & 523 & 327 & 237 & 658 & 535 & 506 & 734 & 403 \\
& Time (s.) & 0.0010 & 0.0026 & 0.0030 & 0.0020 & 0.0029 & 0.0051 & 0.0020 & 0.0030 & 0.0050 & 0.0060 & 0.0046 & 0.0030 & 0.0049 & 0.0051 & 0.0075 \\
\hline
\multirow{3}{*}{4} & Makespan & 41 & 32 & 221 & 74 & 186 & 76 & 161 & 523 & 323 & 241 & 683 & 534 & 471 & 707 & 422 \\
& Balance & 37 & 31 & 220 & 74 & 181 & 64 & 161 & 523 & 307 & 241 & 683 & 534 & 470 & 707 & 422 \\
& Time (s.) & 0.0103 & 0.0269 & 0.0945 & 0.0365 & 0.0470 & 0.1121 & 0.0785 & 0.1628 & 0.2222 & 0.4128 & 0.2016 & 0.1891 & 0.5986 & 0.3930 & 0.7291 \\
\hline
\multirow{3}{*}{5} & Makespan & 41 & 32 & 221 & 74 & 186 & 76 & 161 & 523 & 323 & 241 & 683 & 534 & 471 & 707 & 422 \\
& Balance & 37 & 31 & 220 & 74 & 181 & 64 & 161 & 523 & 307 & 241 & 683 & 534 & 470 & 707 & 422 \\
& Time (s.) & 0.0105 & 0.0263 & 0.1107 & 0.0330 & 0.0674 & 0.0932 & 0.0743 & 0.1749 & 0.2703 & 0.4290 & 0.2146 & 0.2012 & 0.6311 & 0.4117 & 0.6650 \\
\hline
\multirow{3}{*}{6} & Makespan & 41 & 32 & 221 & 74 & 186 & 76 & 161 & 523 & 323 & 241 & 683 & 534 & 471 & 707 & 422 \\
& Balance & 37 & 31 & 220 & 74 & 181 & 64 & 161 & 523 & 307 & 241 & 683 & 534 & 470 & 707 & 422 \\
& Time (s.) & 0.0105 & 0.0239 & 0.1184 & 0.0410 & 0.0501 & 0.0979 & 0.0805 & 0.1539 & 0.2228 & 0.4545 & 0.2256 & 0.1957 & 0.6254 & 0.3922 & 0.6932 \\
\hline
\multirow{3}{*}{7} & Makespan & 41 & 32 & 221 & 74 & 186 & 76 & 161 & 523 & 323 & 241 & 683 & 534 & 471 & 707 & 422 \\
& Balance & 37 & 31 & 220 & 74 & 181 & 64 & 161 & 523 & 307 & 241 & 683 & 534 & 470 & 707 & 422 \\
& Time (s.) & 0.0101 & 0.0233 & 0.1077 & 0.0354 & 0.0518 & 0.1121 & 0.0817 & 0.1718 & 0.2387 & 0.4742 & 0.2033 & 0.2106 & 0.6106 & 0.4129 & 0.7008 \\
\hline
\multirow{3}{*}{8} & Makespan & 49 & 33 & 204 & 73 & 191 & 74 & 166 & 531 & 348 & 249 & 659 & 559 & 519 & 753 & 424 \\
& Balance & 42 & 33 & 204 & 67 & 180 & 59 & 166 & 523 & 299 & 224 & 659 & 559 & 519 & 734 & 378 \\
& Time (s.) & 0.0010 & 0.0010 & 0.0030 & 0.0023 & 0.0011 & 0.0030 & 0.0020 & 0.0030 & 0.0050 & 0.0050 & 0.0021 & 0.0022 & 0.0040 & 0.0061 & 0.0054 \\
\hline
\multirow{3}{*}{9} & Makespan & 43 & 34 & 221 & 74 & 186 & 77 & 164 & 523 & 323 & 225 & 659 & 534 & 484 & 707 & 444 \\
& Balance & 43 & 32 & 220 & 74 & 178 & 70 & 164 & 523 & 307 & 223 & 659 & 534 & 484 & 707 & 442 \\
& Time (s.) & 0.0145 & 0.0322 & 0.1433 & 0.0404 & 0.0554 & 0.1439 & 0.1018 & 0.2264 & 0.3837 & 0.6498 & 0.2422 & 0.2435 & 0.9498 & 0.5214 & 0.9849 \\
\hline
\multirow{3}{*}{10} & Makespan & 41 & 32 & 221 & 74 & 186 & 76 & 161 & 523 & 323 & 241 & 683 & 534 & 471 & 707 & 422 \\
& Balance & 37 & 31 & 220 & 74 & 181 & 64 & 161 & 523 & 307 & 241 & 683 & 534 & 470 & 707 & 422 \\
& Time (s.) & 0.0100 & 0.0267 & 0.1054 & 0.0299 & 0.0412 & 0.1025 & 0.0830 & 0.1687 & 0.2517 & 0.4734 & 0.1986 & 0.1940 & 0.7519 & 0.3301 & 0.6851 \\
\hline
\multirow{3}{*}{11} & Makespan & 41 & 32 & 221 & 74 & 186 & 76 & 161 & 523 & 323 & 241 & 683 & 534 & 471 & 707 & 422 \\
& Balance & 37 & 31 & 220 & 74 & 181 & 64 & 161 & 523 & 307 & 241 & 683 & 534 & 470 & 707 & 422 \\
& Time (s.) & 0.0133 & 0.0374 & 0.1260 & 0.0507 & 0.0424 & 0.1082 & 0.0805 & 0.1710 & 0.2690 & 0.4869 & 0.2058 & 0.1937 & 0.7672 & 0.4124 & 0.7319 \\
\hline
\multirow{3}{*}{12} & Makespan & 49 & 33 & 204 & 73 & 191 & 74 & 166 & 531 & 348 & 249 & 659 & 559 & 519 & 753 & 424 \\
& Balance & 42 & 33 & 204 & 67 & 180 & 59 & 166 & 523 & 299 & 224 & 659 & 559 & 519 & 734 & 378 \\
& Time (s.) & 0.0010 & 0.0036 & 0.0030 & 0.0020 & 0.0020 & 0.0031 & 0.0020 & 0.0035 & 0.0041 & 0.0040 & 0.0028 & 0.0035 & 0.0070 & 0.0040 & 0.0050 \\
\hline
\multirow{3}{*}{13} & Makespan & 43 & 34 & 221 & 74 & 186 & 76 & 164 & 523 & 327 & 228 & 659 & 534 & 484 & 707 & 444 \\
& Balance & 43 & 32 & 220 & 74 & 178 & 63 & 164 & 523 & 307 & 227 & 659 & 534 & 484 & 707 & 442 \\
& Time (s.) & 0.0133 & 0.0359 & 0.1007 & 0.0528 & 0.0506 & 0.1222 & 0.0804 & 0.1697 & 0.2534 & 0.5052 & 0.1815 & 0.2112 & 0.6604 & 0.4210 & 0.7168 \\
\hline
\end{tabular}
}
\end{table*}

\begin{table*}[ht]
\centering
\caption{Results of the Pareto front heuristics generated by REMoH framework on the Brandimarte Instances, via  GPT-4o\cite{openai2023gpt4}.}
\label{tab:gpt_results}
\resizebox{\textwidth}{!}{%
\begin{tabular}{llccccccccccccccccc}
  &  & mk01 & mk02 & mk03 & mk04 & mk05 & mk06 & mk07 & mk08 & mk09 & mk10 & mk11 & mk12 & mk13 & mk14 & mk15 \\ \hline

\multirow{3}{*}{1} & Makespan & 50 & 40 & 232 & 89 & 189 & 95 & 212 & 531 & 369 & 277 & 669 & 563 & 533 & 753 & 435 \\
& Balance & 42 & 38 & 216 & 74 & 183 & 79 & 212 & 523 & 299 & 263 & 669 & 559 & 533 & 734 & 403 \\
& Time (s.) & 0.0010 & 0.0000 & 0.0010 & 0.0000 & 0.0000 & 0.0010 & 0.0010 & 0.0000 & 0.0011 & 0.0020 & 0.0000 & 0.0000 & 0.0010 & 0.0010 & 0.0020 \\
\hline
\multirow{3}{*}{2} & Makespan & 49 & 35 & 204 & 73 & 191 & 81 & 166 & 531 & 349 & 257 & 659 & 559 & 520 & 753 & 434 \\
& Balance & 42 & 34 & 204 & 67 & 180 & 66 & 166 & 523 & 299 & 234 & 659 & 559 & 517 & 734 & 378 \\
& Time (s.) & 0.0000 & 0.0010 & 0.0020 & 0.0000 & 0.0010 & 0.0030 & 0.0010 & 0.0030 & 0.0029 & 0.0050 & 0.0020 & 0.0020 & 0.0029 & 0.0020 & 0.0040 \\
\hline
\multirow{3}{*}{3} & Makespan & 47 & 40 & 219 & 76 & 189 & 99 & 212 & 539 & 366 & 281 & 669 & 560 & 533 & 753 & 441 \\
& Balance & 42 & 38 & 216 & 66 & 183 & 79 & 212 & 523 & 299 & 252 & 669 & 559 & 533 & 734 & 403 \\
& Time (s.) & 0.0010 & 0.0010 & 0.0010 & 0.0000 & 0.0000 & 0.0011 & 0.0010 & 0.0020 & 0.0020 & 0.0020 & 0.0010 & 0.0010 & 0.0010 & 0.0020 & 0.0010 \\
\hline
\multirow{3}{*}{4} & Makespan & 45 & 29 & 204 & 82 & 186 & 80 & 168 & 523 & 351 & 264 & 659 & 559 & 513 & 763 & 426 \\
& Balance & 42 & 29 & 204 & 74 & 184 & 74 & 168 & 523 & 336 & 242 & 659 & 559 & 513 & 734 & 396 \\
& Time (s.) & 0.0000 & 0.0010 & 0.0010 & 0.0010 & 0.0000 & 0.0010 & 0.0010 & 0.0020 & 0.0020 & 0.0020 & 0.0000 & 0.0010 & 0.0040 & 0.0010 & 0.0020 \\
\hline
\multirow{3}{*}{5} & Makespan & 45 & 31 & 204 & 82 & 186 & 79 & 163 & 523 & 352 & 256 & 653 & 540 & 520 & 763 & 430 \\
& Balance & 42 & 30 & 204 & 76 & 184 & 72 & 163 & 523 & 319 & 222 & 653 & 535 & 520 & 734 & 391 \\
& Time (s.) & 0.0010 & 0.0011 & 0.0010 & 0.0010 & 0.0000 & 0.0010 & 0.0010 & 0.0020 & 0.0010 & 0.0020 & 0.0010 & 0.0010 & 0.0020 & 0.0010 & 0.0010 \\
\hline
\multirow{3}{*}{6} & Makespan & 46 & 31 & 204 & 73 & 191 & 73 & 160 & 531 & 340 & 244 & 671 & 559 & 502 & 753 & 441 \\
& Balance & 42 & 29 & 204 & 67 & 181 & 57 & 160 & 523 & 299 & 214 & 671 & 559 & 502 & 734 & 378 \\
& Time (s.) & 0.0010 & 0.0010 & 0.0010 & 0.0010 & 0.0020 & 0.0020 & 0.0000 & 0.0021 & 0.0020 & 0.0037 & 0.0010 & 0.0020 & 0.0020 & 0.0010 & 0.0021 \\
\hline
\multirow{3}{*}{7} & Makespan & 46 & 31 & 204 & 73 & 191 & 73 & 160 & 531 & 340 & 244 & 671 & 559 & 502 & 753 & 441 \\
& Balance & 42 & 29 & 204 & 67 & 181 & 57 & 160 & 523 & 299 & 214 & 671 & 559 & 502 & 734 & 378 \\
& Time (s.) & 0.0000 & 0.0010 & 0.0020 & 0.0010 & 0.0010 & 0.0010 & 0.0020 & 0.0020 & 0.0020 & 0.0031 & 0.0010 & 0.0009 & 0.0030 & 0.0020 & 0.0020 \\
\hline
\multirow{3}{*}{8} & Makespan & 46 & 31 & 207 & 73 & 191 & 76 & 160 & 531 & 344 & 265 & 671 & 559 & 502 & 769 & 433 \\
& Balance & 42 & 29 & 204 & 67 & 181 & 61 & 160 & 523 & 299 & 221 & 671 & 559 & 502 & 734 & 378 \\
& Time (s.) & 0.0010 & 0.0010 & 0.0020 & 0.0010 & 0.0020 & 0.0020 & 0.0010 & 0.0010 & 0.0020 & 0.0030 & 0.0010 & 0.0009 & 0.0020 & 0.0010 & 0.0022 \\
\hline
\multirow{3}{*}{9} & Makespan & 50 & 40 & 226 & 89 & 189 & 95 & 212 & 531 & 369 & 277 & 669 & 563 & 533 & 753 & 449 \\
& Balance & 42 & 38 & 216 & 74 & 183 & 79 & 212 & 523 & 299 & 263 & 669 & 559 & 533 & 734 & 403 \\
& Time (s.) & 0.0000 & 0.0000 & 0.0000 & 0.0000 & 0.0000 & 0.0000 & 0.0000 & 0.0010 & 0.0010 & 0.0010 & 0.0009 & 0.0000 & 0.0010 & 0.0010 & 0.0021 \\

\hline
\end{tabular}
}
\end{table*}

\section{Heuristics generated}

In this section, we provide an overview of the main groups of heuristics generated by \texttt{Gemini 2.0 Flash}, the LLM that achieved the best performance in the models comparison presented in \ref{apendix:llm_selection}. The goal is to better understand the types of heuristics favored by the model when optimizing the FJSSP across different instances.

The final Pareto front from REMoH includes 13 diverse heuristics, each implementing a unique scheduling strategy or parameter configuration. These heuristics were categorized into three main groups based on their core characteristics. These categories are not mutually exclusive since some heuristics integrate elements from multiple groups.

\begin{enumerate}
    \item \textbf{Basic Priority Heuristics with Weight Adjustment}: These heuristics are based on a weighted priority function to select the next operation to schedule. The key is the dynamic adjustment of the weights of different factors (makespan, load, urgency) during the scheduling process.
    \item \textbf{Priority Heuristics with SPT and Load Balancing}: These heuristics combine the Shortest Processing Time (SPT) rule to balance the machine load. They use a cost function considering the start time, processing time, and workload. Some of these heuristics use an iterative approach to improve the solution. 
    \item \textbf{Advanced Heuristics with Enhanced Features}:
    These extend basic strategies by integrating advanced mechanisms like lookahead that considers the impact of the decision on future operations, bottleneck prioritization, dynamic scaling of priority function, and non-linear trade-off functions (e.g., sigmoid) to balance objectives.
\end{enumerate}

\section{Prompts examples}\label{apendix:prompts}

The prompts used in the REMoH algorithm are crucial for understanding and potentially improving the performance of the LLM. In this case, they were mainly adapted from HSEvo, with modifications to tailor the model’s behavior to our specific needs.

In this section, we will explain and analyze the main parts of the prompts used.

\subsection{General prompts}

General prompts refer to instructions reused in different steps of our algorithm. Firstly, the generator prompt serves as a concise instruction delivered to the LLM during each stage of heuristic generation (Prompt 1).

\begin{tcolorbox}[title=Prompt 1: Generator Prompt, colback=gray!5, colframe=yellow!25, coltitle=black, boxrule=0.8pt, arc=4pt]
{\texttt{\{role instruction\}}} helping to design heuristics that can effectively solve optimization problems.

Your response outputs Python code and nothing else. Format your code as a Python code string: 

```python ... ```.
\end{tcolorbox}

As outlined in the methodology \ref{sec: method}, a dynamic prompt \texttt{role instruction} is employed to foster diversity, incorporating multiple roles specifically crafted for the LLM to enhance heuristic development. As shown in Prompt 2, it consists of the name of a famous scientist along with a brief description of their main contributions.

\begin{tcolorbox}[title=Prompt 2: Different role instruction prompts
, colback=gray!5, colframe=yellow!25, coltitle=black, boxrule=0.8pt, arc=4pt]
You are \texttt{\{scientific\}}, \texttt{\{description of their main contributions\}}
\end{tcolorbox}

Additionally, the task description is a key instruction provided to the LLM, as it outlines the details of the optimization problem and the heuristic function to be generated, including the expected input and output formats (Prompt 3).

\begin{tcolorbox}[title=Prompt 3: Task description prompt, colback=gray!5, colframe=yellow!25, coltitle=black, boxrule=0.8pt, arc=4pt]
{\texttt{\{role instruction\}}} Your task is to write a function called 'heuristic' for {\texttt{\{problem description\}}}

{\texttt{\{function description\}}}
\end{tcolorbox}

In the task prompt, \texttt{problem description} and \texttt{function description} depend on the optimization problem to resolve. Prompts used for FJSSP are illustrated in Prompt 4 and 5, respectively.

\begin{tcolorbox}[title=Prompt 4: Problem description prompt for FJSSP, colback=gray!5, colframe=yellow!25, coltitle=black, boxrule=0.8pt, arc=4pt]
The Flexible Job Shop Scheduling Problem (FJSSP) is an optimization problem where jobs, consisting of multiple operations, must be scheduled on machines with specific processing times.
The goal is to minimize makespan, reduce idle time between operations in the same job, and balance workload.
Ensure that the following constraints are strictly followed:
    -Operation feasibility: Each operation must be performed on the corresponding machine(s), and the processing time must align with the available machines.
    -Machine feasibility: Machines can perform only one operation at a time, ensuring no overlap in their schedules.
    -Sequence feasibility: Within the same job, operations must be executed in their defined numerical order. A subsequent operation must only start once the previous one has completed, ensuring proper sequencing and avoiding any premature execution of operations.
\end{tcolorbox}

\begin{tcolorbox}[title=Prompt 5: Function description prompt for FJSSP, colback=gray!5, colframe=yellow!25, coltitle=black, boxrule=0.8pt, arc=4pt]
The input of the function 'heuristic' is a dictionary with the following keys:
- n jobs: Total number of jobs.
- n machines: Total number of machines.
- jobs: A dictionary where:
    - Each key is a job number.
    - Each value in the list represents a set of operations for a job. The index of each operation within the list indicates the order in which they must be executed in the scheduling.
    - Each operation is a tuple containing:
        - A list of machines involved. (start from 0)
        - A list of corresponding processing times.

The output is a dictionary where:
- Each **key** is a **job number**.
- The value is a list of dictionaries containing each operation as:
    - Operation: Operation number.
    - Assigned Machine: Assigned machine number.
    - Start Time: Start time.
    - End Time: End time.
    - Processing Time: Processing time.
\end{tcolorbox}

\subsection{Population initialization prompt}

The initialization prompt assists the model in generating the first heuristics. To achieve this, the instruction incorporates the {\texttt{{generator prompt}}} as shown in Prompt 1. Additionally, it provides the {\texttt{{task description}}} (Prompt 3), and a {\texttt{{seed function}}} is used to enable the model to leverage existing knowledge. It can be read in Prompt 6.

\begin{tcolorbox}[title=Prompt 6: Population initialization prompt, colback=gray!5, colframe=yellow!25, coltitle=black, boxrule=0.8pt, arc=4pt]
{\texttt{\{generator prompt\}}}

{\texttt{\{task description\}}}

{\texttt{\{seed function\}}}

Refer to the format of a trivial design above. 

Be very creative and give a completely different heuristic that improves the one given, respeting constraints. 

Output code only and enclose your code with Python code block: ´´´python ... ´´´,  has comment and docstring (<50 words) to description key idea of heuristics design.


Let's think step by step.
\end{tcolorbox}

\subsection{Crossover prompt}

To improve the generation of new heuristics based on parents and reflection, a crossover prompt is employed. As shown in Prompt 7, it includes the {\texttt{{generator prompt}}} and {\texttt{{task description}}} previously explained. Additionally, the code for both {\texttt{{parent1}}} and {\texttt{{parent2}}} is provided, along with the {\texttt{{long reflection}}}.

\begin{tcolorbox}[title=Prompt 7: Crossover prompt, colback=gray!5, colframe=yellow!25, coltitle=black, boxrule=0.8pt, arc=4pt, label={box:crossoverprompt}]
{\texttt{\{generator prompt}\}}

{\texttt{\{task description}\}}

\#\#\# Parent 1

{\texttt{\{parent1}\}}

\#\#\# Parent 2

{\texttt{\{parent2}\}}

\#\#\# Analyze \& experience

- {\texttt{\{long reflection}\}}

Your task is to write an improved function 
named heuristic by COMBINING elements of
two above heuristics base Analyze \& experience.
Output the code within a Python code block:
'''python ... ''', has comment and docstring ($<50$
words) to description key idea of heuristics design.
Let’s think step by step.
\end{tcolorbox}

\subsection{Elitist mutation prompt}

The elitist mutation prompt aids the LLM in generating new heuristics based on the best existing one. It consists of the {\texttt{{generator prompt}}} and {\texttt{{task description}}}, with the addition of the code for the best heuristic {\texttt{{parent}}} and the {\texttt{{long reflection}}}. It is illustrated in Prompt 8.

\begin{tcolorbox}[title=Prompt 8: Elitist mutation prompt, colback=gray!5, colframe=yellow!25, coltitle=black, boxrule=0.8pt, arc=4pt, label={box:mutationprompt}]
{\texttt{\{generator prompt}\}}

{\texttt{\{task description}\}}

Current heuristics:

{\texttt{\{parent}\}}

Now, think outside the box and write a mutated function better than current version. 

You can using some hints if need:

{\texttt{\{long reflection}\}}

Output code only and enclose your code with Python code block: ´´´python ... ´´´, has comment and docstring ($<50$ words) to description key idea of heuristics design.

Let's think step by step.
\end{tcolorbox}

\subsection{Reflections prompts}

The reflection prompts are crucial for the LLM to understand what to extract from each cluster and how to combine the short reflections to generate a useful analysis for the creation of new heuristics.

The short reflection (Prompt 9) is provided with the cluster centroid performance on each objective function, stored in  \texttt{cluster performance}. Furthermore, all heuristics code are added in \texttt{heuristics}
\begin{tcolorbox}[title=Prompt 9: Short reflection prompt, colback=gray!5, colframe=yellow!25, coltitle=black, boxrule=0.8pt, arc=4pt]
You are an expert in heuristics for multiobjective problem optimization. You will be given a cluster of heuristics. It contains heuristics that have similar performance across different objective functions. Your task is to analyze the heuristics and summarize their characteristics and trends, highlighting the key features with no more than 50 words. Avoid redundant information.

The general cluster performance is {\texttt{\{cluster performance}\}}

Scores are scaled using a Standard Scaler. Positive values indicate performance above the mean (better), while negative values indicate performance below the mean (worse).


Let's think step by step.

\#\#\#Heuristics

{\texttt{\{heuristics}\}}
\end{tcolorbox}

The long reflection, shown in Prompt 10, includes the previous long reflection (\texttt{long reflection}) as well as the short reflections extracted with Prompt 9, each accompanied by the performance of the corresponding cluster centroid in \texttt{cluster reflections}.

\begin{tcolorbox}[title=Prompt 10: Long reflection prompt, colback=gray!5, colframe=yellow!25, coltitle=black, boxrule=0.8pt, arc=4pt]
You are an expert in the domain of optimization heuristics. Your task is to provide useful advice based on analysis to design better heuristics.  Answer just what is asked and avoid redundant information.

Here are previous long-term reflections on heuristic performance:

\#\# Previous Long-Term Reflection:

{\texttt{\{long reflection}\}}

Now, analyze the following cluster reflections:

\#\# Clusters Reflections:

{\texttt{\{cluster reflections}\}}

Scores are scaled using a Standard Scaler. Positive values indicate performance above the mean (better), while negative values indicate performance below the mean (worse).

Synthesize a general reflection that identifies overarching trends, highlights major strengths and weaknesses and connects insights from both the clusters and the long-term perspective. Your reflection should guide LLMs in designing more effective heuristics that optimize all objective functions.  Limit your response to 150 words.


Let's think step by step.
\end{tcolorbox}

\bibliographystyle{IEEEtran}
\bibliography{cas-refs}

\newpage

\section{Biography Section}

\vspace{11pt}

\vspace{-33pt}
\begin{IEEEbiography}[{\includegraphics[width=1in,height=1.25in,clip,keepaspectratio]{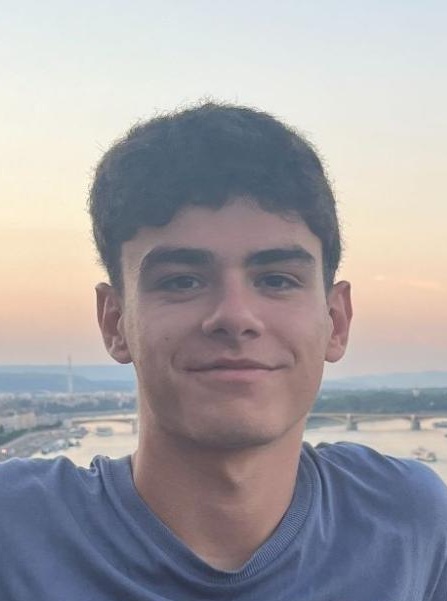}}]{Diego Forniés-Tabuenca} is a final-year student in the bachelor's degree in Artificial Intelligence at the Public University of the Basque Country. He is currently an intern at the Foundation Center for Visual Interaction and Communications Technologies, Vicomtech, working in the Data Intelligence for Energy and Industrial Processes department. He is interested in the application of artificial intelligence in natural language processing, industry, and biomedicine.
\end{IEEEbiography}

\vspace{-33pt}
\begin{IEEEbiography}[{\includegraphics[width=1in,height=1.25in,clip,keepaspectratio]{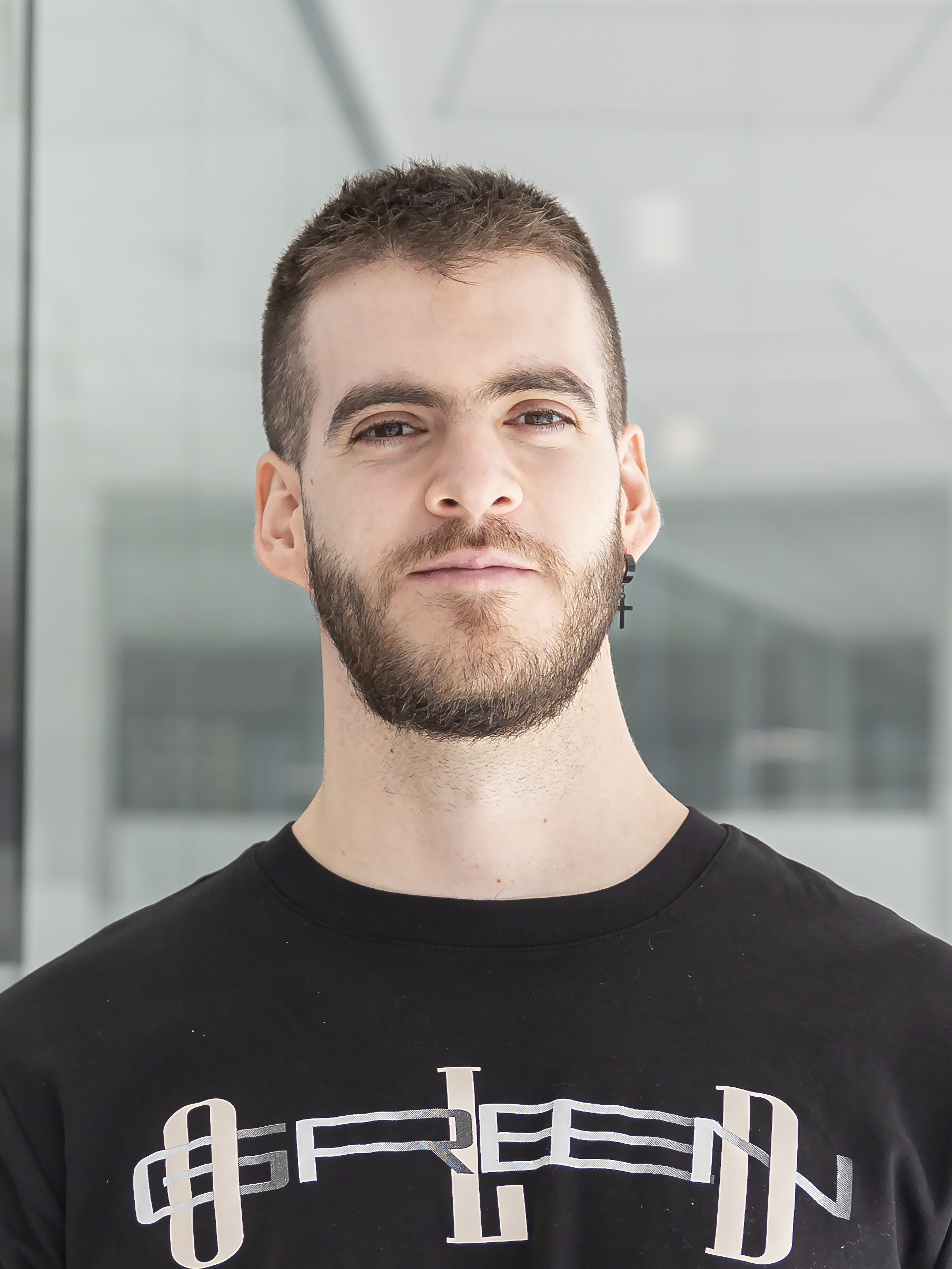}}]{Alejandro Uribe} has a bachelor’s degree in Physics Engineering and a Master’s degree in Engineering from EAFIT University. He is currently studying Mathematical Engineering PhD at the same university and working as a research assistant at the  Foundation Center for Visual Interaction and Communications Technologies, Vicomtech. His studies have focused on simulation and optimization using Python programming tools. He is interested in mathematical problem formulation, heuristics, artificial intelligence, and new technologies.
\end{IEEEbiography}

\vspace{-33pt}
\begin{IEEEbiography}[{\includegraphics[width=1in,height=1.25in,clip,keepaspectratio]{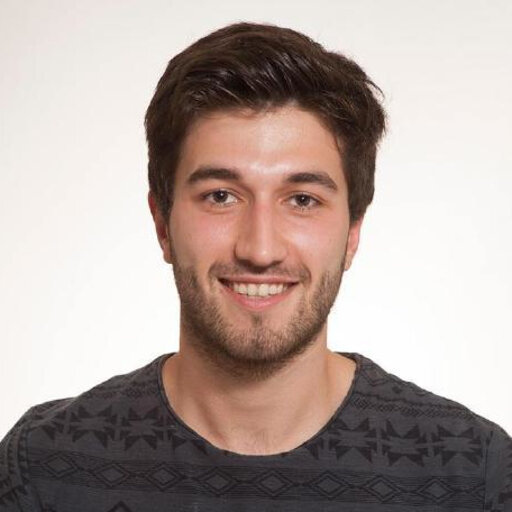}}]{Urtzi Otamendi} has a bachelor’s degree in Computer Engineering from the Faculty of Computer Science of the Public University of the Basque Country. He completed his Master’s Degree in Artificial Intelligence at the Polytechnic University of Madrid. He works as a research assistant in the Foundation Center for Visual Interaction and Communications Technologies, Vicomtech, in the Data Intelligence for Energy and Industrial Processes department.
His interests include software development, artificial intelligence, and computer vision applied to innovative technologies.
\end{IEEEbiography}

\vspace{-33pt}
\begin{IEEEbiography}[{\includegraphics[width=1in,height=1.25in,clip,keepaspectratio]{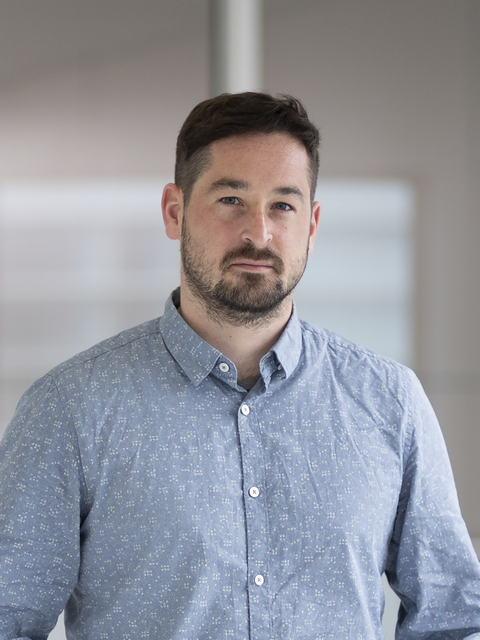}}]{Arkaitz Artetxe} holds a PhD in Computer Science and a Master’s in Computational Engineering and Intelligent
Systems, both from the University of the Basque Country. He is a researcher at Vicomtech, where he has worked in e-health and biomedical
applications as well as in data intelligence for energy and industrial processes. His work focuses on applying computational engineering and
data mining to real-world applications. He has published and reviewed extensively in high-impact international conferences and journals.
\end{IEEEbiography}

\vspace{-33pt}
\begin{IEEEbiography}[{\includegraphics[width=1in,height=1.25in,clip,keepaspectratio]{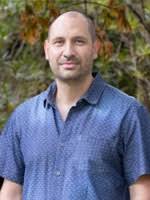}}]{Juan Carlos Rivera} holds a Bachelor’s degree in Industrial Engineering and a Master’s in Computer Science from the National University of Colombia (Medellin campus). He earned a Ph.D. in System Optimization and Reliability from the University of Technology of Troyes (France). Currently, he is a Full Professor at EAFIT University in the field of Computing and Analytics. His main research interests focus on the development of mixed-integer linear
programming models and metaheuristic strategies to solve optimization problems, with applications in logistics and production.
\end{IEEEbiography}

\vspace{-33pt}
\begin{IEEEbiography}[{\includegraphics[width=1in,height=1.25in,clip,keepaspectratio]{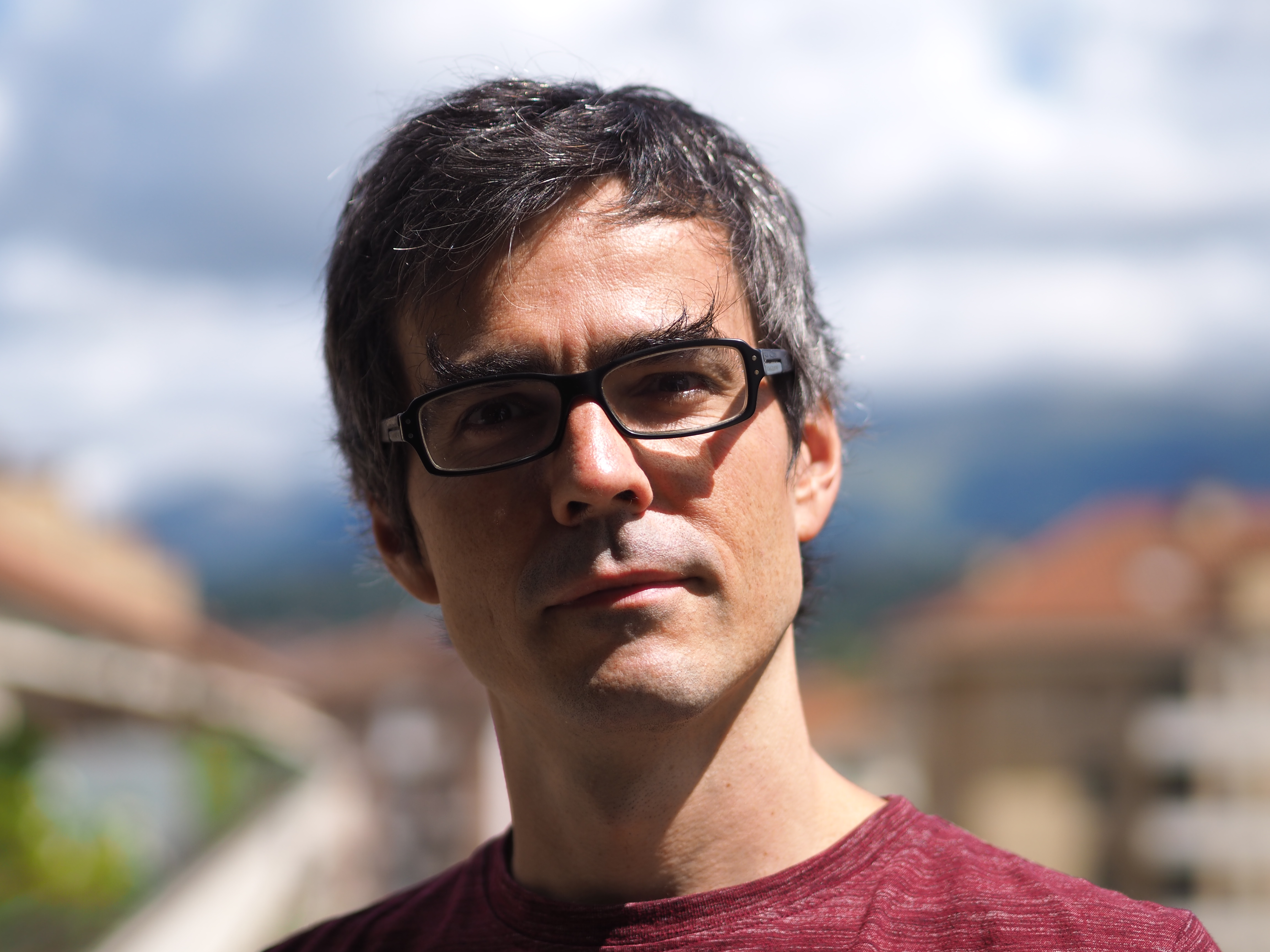}}]{Oier Lopez de Lacalle} is an Assistant Professor at the University of the Basque Country and an adjunct Senior Researcher at the Basque Center for Language Technology (HiTZ), Spain. He received his PhD in Natural Language Processing and Computer Science from the University of the Basque Country. His research interests include language modelling for under-resourced languages, multimodal learning, and grounded representations. He has published and reviewed extensively in high-impact international conferences and journals.
\end{IEEEbiography}

\vspace{11pt}

\vfill

\end{document}